# CTSR: Cartesian tensor-based sparse regression for data-driven discovery of high-dimensional invariant governing equations

Boqian Zhang, Juanmian Lei *, Guoyou Sun, Shuaibing Ding and Jian Guo

School of Aerospace Engineering, Beijing Institute of Technology, Beijing, 100081, China

Accurate and concise governing equations are crucial for understanding system dynamics. Recently, data-driven methods such as sparse regression have been employed to automatically uncover governing equations from data, representing a significant shift from traditional first-principles modeling. However, most existing methods focus on scalar equations, limiting their applicability to simple, low-dimensional scenarios, and failing to ensure rotation and reflection invariance without incurring significant computational cost or requiring additional prior knowledge. This paper proposes a Cartesian tensor-based sparse regression (CTSR) technique to accurately and efficiently uncover complex, high-dimensional governing equations while ensuring invariance. Evaluations on two two-dimensional (2D) and two three-dimensional (3D) test cases demonstrate that the proposed method achieves superior accuracy and efficiency compared to the conventional technique.

## I. Introduction

Many physical phenomena in nature are described by governing equations, typically formulated as partial differential equations. Accurate and concise representations of these equations not only deepen our understanding of the underlying mechanisms but also provide the mathematical basis for analytical and numerical analyses of physical problems. Traditionally, governing equations—such as the Navier–Stokes equation for fluid motion—are derived from first principles like conservation laws. However, this approach is significantly limited by the researcher's expertise and the system's complexity. In highly complex dynamic systems, such as multiphase flows or meteorological processes, the high dimensionality and numerous parameters render derivations based solely on first principles exceedingly challenging.

* E-mail: leijm@bit.edu.cn

Recent advances in machine learning and the growing availability of observational data, experimental measurements, and numerical simulations have established data-driven methods as a novel paradigm for discovering governing equations [1,2]. One representative class comprises evolutionary algorithm-based approaches [3–12], such as Eureqa [7], SGA-PDE [9], and PySR [11]. These methods construct a symbolic library of physical variables and mathematical operators and then search for appropriate symbol combinations to generate target equations. Owing to their minimal restrictions on target equation forms, these methods offer high flexibility and expressivity. However, they are sensitive to hyperparameter settings and, without suitable constraints, tend to be computationally inefficient and challenging to scale to high-dimensional datasets [2]. Another category relies on neural networks [13–20], exemplified by EQL[14], PDE-NET[15], and DSR[18]. These techniques capitalize on the strong approximation capabilities of neural networks and enable efficient derivative computation via backpropagation. Their drawbacks include high training costs for complex problems and representational limitations imposed by network architectures. Additionally, the use of operations such as logarithms and exponentials can introduce numerical issues [2] or substantially reduce computational efficiency.

Sparse regression has emerged as another widely studied approach for equation discovery. This technique assumes that a target equation can be expressed as a linear combination of an overcomplete set of nonlinear candidate functions, thereby recasting the discovery problem as linear regression and employing sparsity-promoting techniques to achieve a parsimonious representation. Brunton et al. [21] introduced the sparse identification of nonlinear dynamics (SINDy) method for extracting ordinary differential equations from data, while Rudy et al. [22] proposed the PDE-FIND method, which extends this framework to the discovery of partial differential equations. Further refinements, including techniques that incorporate weak formulations to mitigate the effects of noisy data, have been proposed [23,24]. Additional related studies are provided in Ref. [25–30]. Although assuming a specific form for the governing equations constrains the expressive capabilities of sparse regression methods, its computational efficiency is far superior to that of evolutionary algorithms and neural network-based approaches, and facilitates the derivation of concise governing equations.

Despite the strong potential of data-driven approaches for discovering governing equations, existing methods face two key challenges. First, most real-world physical systems exhibit invariance under coordinate transformations, such as translations, rotations, and reflections. Noether's theorem establishes a link between invariance and conservation [31], highlighting the importance of invariance in formulating governing equations. However, a unified framework for uncovering invariant governing equations from data remains absent. Existing studies typically impose invariance constraints by either selecting or constructing invariant candidate functions [24,32], verifying the invariance of predictions [16], or applying data augmentation techniques [33]. The first strategy relies heavily on prior knowledge and manual intervention, limiting its extension to more complex or higher-dimensional problems, whereas the latter two approaches reduce this reliance but incur significantly higher computational costs.

The second challenge arises from the fact that most existing methods are designed for scalar equations. Consequently, as both the dimensionality and order of the equations increase, the number of candidate functions and equations grows substantially. In complex, high-dimensional dynamical systems, this results in a dramatic expansion of the parameter search space, escalating both the learning difficulties and computational burdens. As a result, without substantially enhancing computational resources or relying on extensive prior knowledge, current algorithms are predominantly applied to relatively simple, low-dimensional problems [22,34] or to reduced-order models of high-dimensional systems [35]. Although a few methods have been developed for tensor equations—such as the M-GEP method proposed by Weatheritt and Sandberg [36], which uses hosts and plasmids to represent tensors and their scalar coefficients and employs Cayley-Hamilton theory to construct tensor bases as invariant input features—the construction of tensor bases depends on complex derivations and additional assumptions, and the method's applicability beyond turbulence modeling remains unverified. Therefore, there is a clear need for an efficient and universal framework capable of uncovering invariant governing equations for complex, high-dimensional dynamical systems with minimal reliance on physical priors.

This paper introduces a Cartesian tensor-based sparse regression (CTSR) approach. By formulating the target equations in Cartesian tensor form, the proposed method inherently satisfies rotation and reflection invariance without incurring significant computational overhead or requiring additional physical priors. The CTSR approach utilizes the TrainSTRidge sparse regression algorithm to determine the equation coefficients, while Pareto analysis is employed to select key hyperparameters. The method's effectiveness is demonstrated through four examples of increasing complexity and dimensionality: the 2D Burgers equation, the 2D Navier–Stokes equation, the 3D Navier–Stokes equation, and the 3D Giesekus equation. In addition, this study examines the role of Cartesian tensors in the equation discovery process, evaluates the influence of sampling points on prediction errors, and assesses the algorithm's runtime performance.

The remainder of the paper is organized as follows: Section II introduces Cartesian tensor equations, providing mathematical derivations to demonstrate their invariance under coordinate rotations and reflections. This section also details the construction of the tensor candidate library, the sparse regression algorithm, and the selection of hyperparameters. Section III evaluates the predictive accuracy of the proposed method and compares it with the existing method. Section IV discusses the role of Cartesian tensors, the effect of sampling points, and the algorithm's runtime performance. Finally, Section V concludes the study.

## II. Methodology

### A. Cartesian tensor equation

Tensors serve as fundamental algebraic tools in physics. When defined in a Cartesian coordinate system, they are referred to as Cartesian tensors. In Euclidean 3-space $\mathbb{E}^3$, a zeroth-order Cartesian tensor (e.g., pressure) is a scalar, while a first-order Cartesian tensor (e.g., velocity) has three components. It can be expressed as $\boldsymbol{a}=a_1\boldsymbol{e}_1+a_2\boldsymbol{e}_2+a_3\boldsymbol{e}_3$, where $\{\boldsymbol{e}_1, \boldsymbol{e}_2, \boldsymbol{e}_3\}$ is an orthogonal set of basis vectors and $a_1$, $a_2$, $a_3$ are the components along these vectors. According to the Einstein summation convention, any suffix that appears twice in a single term (i.e., the repeated suffix) is summed over, whereas a suffix that appears only once (i.e., the free suffix) is independent and can take on values 1, 2, or 3. Thus, a first-order tensor can also be concisely written as $\boldsymbol{a}=a_i\boldsymbol{e}_i$, or simply as $a_i$ when the basis is understood. Similarly, a second-order Cartesian tensor (e.g., stress) consists of nine components and can be represented as $\boldsymbol{a}=a_{ij}\boldsymbol{e}_i\boldsymbol{e}_j$, or abbreviated as $a_{ij}$. Representations for higher-order Cartesian tensors can be derived in an analogous manner. Furthermore, multiple Cartesian tensors can be combined to form Cartesian tensor expressions through operations such as contraction and derivative, and these expressions can be linked via the operators "+", "–", and "=" to formulate Cartesian tensor equations.

Most dynamical systems—such as those encountered in fluid flow, heat transfer, and electromagnetism—have governing equations that can be succinctly expressed in the form of Cartesian tensor equations. For example, the nondimensional 3D Navier–Stokes equation, when written in scalar form, is given by,

$$
\begin{cases}
\dfrac{\partial u}{\partial t} = -u\dfrac{\partial u}{\partial x} - v\dfrac{\partial u}{\partial y} - w\dfrac{\partial u}{\partial z} + \dfrac{1}{\mathrm{Re}}\left(\dfrac{\partial^2 u}{\partial x \partial x} + \dfrac{\partial^2 u}{\partial y \partial y} + \dfrac{\partial^2 u}{\partial z \partial z}\right) - \dfrac{\partial p}{\partial x} \\
\dfrac{\partial v}{\partial t} = -u\dfrac{\partial v}{\partial x} - v\dfrac{\partial v}{\partial y} - w\dfrac{\partial v}{\partial z} + \dfrac{1}{\mathrm{Re}}\left(\dfrac{\partial^2 v}{\partial x \partial x} + \dfrac{\partial^2 v}{\partial y \partial y} + \dfrac{\partial^2 v}{\partial z \partial z}\right) - \dfrac{\partial p}{\partial y} \\
\dfrac{\partial w}{\partial t} = -u\dfrac{\partial w}{\partial x} - v\dfrac{\partial w}{\partial y} - w\dfrac{\partial w}{\partial z} + \dfrac{1}{\mathrm{Re}}\left(\dfrac{\partial^2 w}{\partial x \partial x} + \dfrac{\partial^2 w}{\partial y \partial y} + \dfrac{\partial^2 w}{\partial z \partial z}\right) - \dfrac{\partial p}{\partial z}
\end{cases}
\tag{1}
$$

While in Cartesian tensor form, the equation is expressed as

$$
\frac{\partial u_i}{\partial t} = -u_j \frac{\partial u_i}{\partial x_j} + \frac{1}{\mathrm{Re}} \frac{\partial^2 u_i}{\partial x_j \partial x_j} - \frac{\partial p}{\partial x_i}
\tag{2}
$$

In Eq. (1), $t$ denotes time; $x$, $y$, and $z$ represent the Cartesian coordinates; $u$, $v$, and $w$ are the corresponding velocity components; $p$ denotes the pressure; and Re denotes the Reynolds number. In Eq. (2), $u_i$ represents the velocity tensor, and $x_i$ denotes the spatial coordinates. Clearly, Eq. (1) is more complex than Eq. (2). For data-driven methods, the additional complexity translates into a larger parameter search space, thereby increasing the challenges of discovering the equation. Moreover, accurately reconstructing Eq. (1) from data requires scalar-based approaches to predict three separate component equations, further elevating both the difficulty and computational cost. In contrast, Eq. (2) maintains a consistent form across all dimensions, thereby enhancing efficiency and facilitating the utilization of data from different orientations.

Another advantage of expressing the governing equations in Cartesian tensor form is their inherent invariance under rotations and reflections. Cartesian tensor equations are constructed through tensor contraction, derivative, and linear combinations. As demonstrated in Ref. [37], these operations, when applied to Cartesian tensors of any order in $\mathbb{E}^3$, yield results that are equivariant under any rotation or reflection transformation. The proof is provided below:

In $\mathbb{E}^3$, let $T_1$ be a Cartesian tensor of order $x$ and $T_2$ a Cartesian tensor of order $y$, sharing $z$ common suffixes. Their contraction $f(T_1,T_2)_{a_1\cdots a_{x-z}b_1\cdots b_{y-z}} = T_{1,\,a_1\cdots a_{x-z}c_1\cdots c_z} \cdot T_{2,\,b_1\cdots b_{y-z}c_1\cdots c_z}$ satisfies the following property under any rotation or reflection characterized by an orthogonal matrix $R$:

$$
\begin{aligned}
f(T_1,T_2)'_{a_1\cdots a_{x-z}b_1\cdots b_{y-z}} &= T'_{1,\,a_1\cdots a_{x-z}c_1\cdots c_z} \cdot T'_{2\,c_1\cdots c_z b_1\cdots b_{y-z}} \\
&= R_{a_1 d_1}\cdots R_{a_{x-z}d_{x-z}} \cdot R_{c_1 f_1}\cdots R_{c_z f_z} \cdot T_{1,\,d_1\cdots d_{x-z}f_1\cdots f_z} \cdot R_{c_1 g_1}\cdots R_{c_z g_z} \cdot R_{b_1 e_1}\cdots R_{b_{y-z}e_{y-z}} \cdot T_{2,\,g_1\cdots g_z e_1\cdots e_{y-z}} \\
&= R_{a_1 d_1}\cdots R_{a_{x-z}d_{x-z}} \cdot R_{b_1 e_1}\cdots R_{b_{y-z}e_{y-z}} \cdot T_{1,\,d_1\cdots d_{x-z}f_1\cdots f_z} T_{2,\,g_1\cdots g_z e_1\cdots e_{y-z}} \cdot \delta_{f_1 g_1}\cdots\delta_{f_z g_z} \\
&= R_{a_1 d_1}\cdots R_{a_{x-z}d_{x-z}} \cdot R_{b_1 e_1}\cdots R_{b_{y-z}e_{y-z}} \cdot T_{1,\,d_1\cdots d_{x-z}f_1\cdots f_z} T_{2,\,f_1\cdots f_z e_1\cdots e_{y-z}} \\
&= R_{a_1 d_1}\cdots R_{a_{x-z}d_{x-z}} \cdot R_{b_1 e_1}\cdots R_{b_{y-z}e_{y-z}} \cdot f(T_1,T_2)_{d_1\cdots d_{x-z}e_1\cdots e_{y-z}}
\end{aligned}
\quad . \tag{3}
$$

where $\delta$ denotes the Kronecker delta.

Similarly, consider the derivative operation $f(T) = \partial / \partial T_{j_1 j_2 \cdots j_n}$, we have:

$$
\begin{aligned}
f(T)' &= \frac{\partial}{\partial T'_{i_1 i_2 \cdots i_n}} \\
&= \frac{\partial}{\partial T_{j_1 j_2 \cdots j_n}} \cdot \frac{\partial T_{j_1 j_2 \cdots j_n}}{\partial T'_{i_1 i_2 \cdots i_n}} \\
&= \frac{\partial}{\partial T_{j_1 j_2 \cdots j_n}} \cdot \frac{\partial T_{j_1 j_2 \cdots j_n}}{R_{j_1 i_1} R_{j_2 i_2} \cdots R_{j_n i_n} \cdot \partial T_{j_1 j_2 \cdots j_n}} \\
&= R_{i_1 j_1} R_{i_2 j_2} \cdots R_{i_n j_n} \frac{\partial}{\partial T_{j_1 j_2 \cdots j_n}} \\
&= R_{i_1 j_1} R_{i_2 j_2} \cdots R_{i_n j_n} f(T)
\end{aligned} \tag{4}
$$

For a linear combination of $m$ $n$-th order Cartesian tensors $f(T_1, T_2, \cdots, T_m) = \sum_{x=1}^{m} C_x T_x$ , we have:

$$
\begin{aligned}
f(T_1, T_2, \cdots, T_m)'_{i_1 i_2 \cdots i_n} &= \sum_{x=1}^{m} C_x T'_{x,\, i_1 i_2 \cdots i_n} \\
&= \sum_{x=1}^{m} C_x R_{i_1 j_1} R_{i_2 j_2} \cdots R_{i_n j_n} \cdot T_{x,\, j_1 j_2 \cdots j_n} \\
&= R_{i_1 j_1} R_{i_2 j_2} \cdots R_{i_n j_n} \cdot f(T_1, T_2, \cdots, T_l)_{j_1 j_2 \cdots j_n}
\end{aligned} \tag{5}
$$

From Eq. (3)-(5), it follows that any Cartesian tensor $T$ in Euclidean 3-space, when subjected to contraction, derivative, or linear combination—denoted collectively as $f(T)$—exhibits equivariance:

$$
f(RT) = Rf(T) \tag{6}
$$

Furthermore, considering the Cartesian tensor equation $g(T)$=0, where $g(T)$ is constructed by the operations of contraction, derivative, and linear combination. Since each operation maintains equivariance, $g(T)$ also satisfies Eq. (6), and thus we have,

$$
g(RT) = Rg\left(T\right) = 0 \tag{7}
$$

Eq. (7) confirms that the form of a Cartesian tensor equation is invariant under rotation or reflection transformation. In classical physics, such invariance is fundamental to most dynamic systems and provides a key principle for deriving or modeling their governing equations. Furthermore, if these governing equations are formulated solely in terms of translation-invariant physical quantities—such as velocity, pressure, or temperature—they will also exhibit translation invariance.

The discussion above indicates that, for high-dimensional dynamic systems adhering to invariance principles, employing Cartesian tensor equations as the target formulation for data-driven equation discovery offers significant advantages over scalar formulations.

## B. Cartesian tensor-based sparse regression

Within the sparse regression framework, the target equation is expressed as a linear combination of an overcomplete set of nonlinear candidate functions. This formulation transforms the equation discovery task into solving the following linear regression problem:

$$\mathbf{u}_t = \mathbf{\Theta}\boldsymbol{\xi} \tag{8}$$

where $\mathbf{u}_t$ denotes the temporal derivative term on the left-hand side of the equation; $\mathbf{\Theta}$ is the candidate library matrix where each column corresponds to a candidate function and each row corresponds to a data point; and $\boldsymbol{\xi}$ is the vector of candidate coefficients to be determined.

Consequently, the equation discovery process involves three main steps: generating the dataset, constructing the matrix $\mathbf{\Theta}$, and solving for the vector $\boldsymbol{\xi}$. These steps are detailed in this subsection.

### 1. Algorithm framework

For discovering governing equations in high-dimensional dynamical systems with invariance properties, we propose a Cartesian tensor-based sparse regression technique, termed CTSR. Its overall framework is illustrated in Fig. 1. Using the Navier–Stokes equation as an example, CTSR comprises the following three main steps:

(a) First, a high-fidelity dataset containing system-related physical quantities (e.g., velocity and pressure) is generated from experiments, observations, or numerical simulations. The spatial and temporal derivatives of these quantities are then computed for the purpose of constructing the candidate library.

(b) Next, the candidate library is constructed through the following processes: input tensor selection, combination, suffixes assignment, filtering and reordering. The library is then integrated with a subsampled dataset from step (a), and the matrix $\mathbf{\Theta}$ is obtained via tensor contraction.

(c) Finally, the vector $\boldsymbol{\xi}$ is determined by applying sparse regression using the TrainSTRidge algorithm.

In this study, the high-fidelity dataset in Step (a) is acquired through numerical simulations, with derivatives computed using the finite difference method. Steps (b) and (c) are described in detail in the following parts.

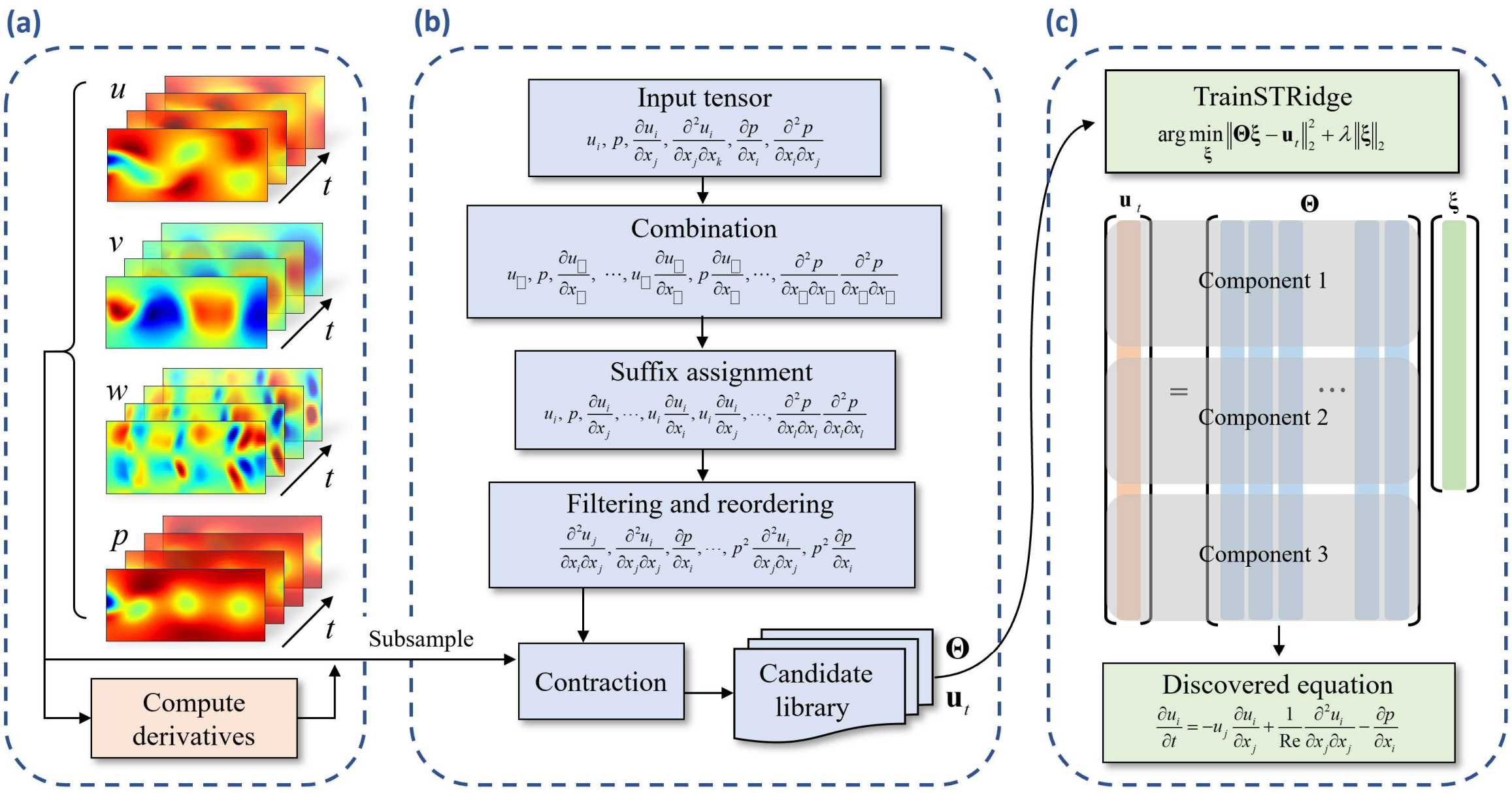

**Fig. 1.** Overall framework of the CTSR. (a) Generate dataset; (b) Construct tensor candidate library; (c) Sparse regression.

## 2. Construction of tensor candidate library

The construction of the tensor candidate library is the most critical and challenging component of CTSR. An appropriate and concise candidate library is essential for accurately discovering the governing equations. This construction process is governed by two fundamental principles: (1) candidate terms must adhere to Cartesian tensor notation rules to ensure *validity*; (2) candidate terms should be mutually independent, free from redundancy and symmetry, to ensure *uniqueness*.

Guided by these principles, we proposed an algorithm for constructing the candidate library, as illustrated in Fig. 2(a). The workflow consists of six steps:

### i. Input tensor selection

Select the input tensors associated with the dynamical system. For example, in the Navier–Stokes equation, we choose the velocity $u_i$, the pressure $p$, and their spatial derivatives up to second order (i.e., $\partial u_i/\partial x_j, \partial^2 u_i/(\partial x_j \partial x_k), \partial p/\partial x_i, \partial^2 p/(\partial x_i \partial x_j)$).

### ii. Combination

Inspired by Ref. [22], partition the input tensors into non-derivative tensors (i.e., $u_i$ and $p$) and derivative tensors (i.e., $\partial u_i/\partial x_j, \partial^2 u_i/(\partial x_j \partial x_k), \partial p/\partial x_i, \partial^2 p/(\partial x_i \partial x_j)$), then represent these terms in Cartesian notation without suffixes (e.g., $u_i$ as $u_{\square}$, and $\partial u_i / \partial x_j$ as $\partial u_{\square} / \partial x_{\square}$). Next, generate all unordered combinations of the non-derivative terms up to a prescribed order $P$. Then, form unordered combinations with each derivative term and constant "1" to generate templates for candidates. For instance, for $P = 1$, the combinations are given by,

$$\begin{aligned} &\{u_{\square}, p, \frac{\partial u_{\square}}{\partial x_{\square}}, \frac{\partial p}{\partial x_{\square}}, \frac{\partial^2 u_{\square}}{\partial x_{\square}\partial x_{\square}}, \frac{\partial^2 p}{\partial x_{\square}\partial x_{\square}}, u_{\square}\frac{\partial u_{\square}}{\partial x_{\square}}, u_{\square}\frac{\partial^2 u_{\square}}{\partial x_{\square}\partial x_{\square}}, \\ &u_{\square}\frac{\partial p}{\partial x_{\square}}, u_{\square}\frac{\partial^2 p}{\partial x_{\square}\partial x_{\square}}, p\frac{\partial u_{\square}}{\partial x_{\square}}, p\frac{\partial^2 u_{\square}}{\partial x_{\square}\partial x_{\square}}, p\frac{\partial p}{\partial x_{\square}}, p\frac{\partial^2 p}{\partial x_{\square}\partial x_{\square}}\} \end{aligned} \tag{9}$$

At this stage, the combinations are symbolic juxtapositions rather than mathematically valid tensor expressions, as suffixes have not yet been assigned.

### iii. Suffix assignment

For each combination obtained from step ii, assign suffixes to form tensor candidates. Specifically, if a combination has total tensor order $n$, there are $n$ positions to assign with $n$ suffixes, yielding $n^n$ distinct schemes. For example, the combination $\partial u_{\square} / \partial x_{\square}$ produces $2^2$=4 possible suffix assignments—namely, $\partial u_i / \partial x_i$, $\partial u_i / \partial x_j$, $\partial u_j / \partial x_i$, and $\partial u_j / \partial x_j$; Similarly, the combination $u_{\square}\partial^2 u_{\square} / (\partial x_{\square}\partial x_{\square})$ leads to $4^4$=64 distinct schemes, such as $u_i \partial^2 u_i / (\partial x_i \partial x_i)$, $u_i \partial^2 u_i / (\partial x_i \partial x_j)$, …, $u_l \partial^2 u_l / (\partial x_l \partial x_k)$, and $u_l \partial^2 u_l / (\partial x_l \partial x_l)$.

It is important to note that most candidates generated at this stage are either invalid, duplicate, or equivalent, and therefore require further filtering.

### iv. Filtering and reordering

In this step, candidates that do not meet Cartesian tensor notation [38] are removed first. Specifically, a candidate is discarded if either of the following conditions is met:

(a) a candidate where the same suffix appears more than twice, e.g., $u_i \partial^2 u_i / (\partial x_i \partial x_i)$;

(b) a candidate whose number of free suffixes differs from the order of the target equation. For example, since the Navier–Stokes equation is considered first order (as $\mathbf{u}_t$ is a first-order tensor), any candidate without exactly one free suffix is invalid.

Additionally, some candidates obtained in step iii are mathematically valid but do not conform to conventional suffix-naming practices. For example, although $\partial u_j / \partial x_j$ can be valid, it is normally written as $\partial u_i / \partial x_i$; similarly, $u_i \partial^2 u_j / (\partial x_l \partial x_l)$ is typically expressed as $u_i \partial^2 u_j / (\partial x_k \partial x_k)$. To standardize, all suffixes are rearranged from left to right and top to bottom in ascending order ($i \rightarrow j \rightarrow k \rightarrow \cdots$). Free suffixes are ordered first, followed by repeated suffixes, while preserving the relative order of the free suffixes. For instance, in $u_l \partial^2 u_k / (\partial x_l \partial x_i)$, the initially suffix set $\{l, k, l, i\}$—with $k$ and $i$ being free suffixes and $l$ being a repeated suffix. In this case, the free suffixes are first reordered (i.e., $k \rightarrow j$ and $i$ remains unchanged), followed by reordering the repeated suffix (i.e., $l \rightarrow k$), yielding the standardized form $u_k \partial^2 u_j / (\partial x_k \partial x_i)$. Furthermore, to ensure uniqueness, if a candidate features symmetric dimensions, the suffixes on these dimensions are reordered in ascending order. For example, in $u_k \partial^2 u_j / (\partial x_k \partial x_i)$, the symmetric pair ($k$, $i$) is reordered to yield $u_k \partial^2 u_j / (\partial x_i \partial x_k)$.

Finally, duplicate or equivalent candidates are removed. These include:

(a) Duplicate candidates. For example, when $\partial u_j / \partial x_j$ is reordered to $\partial u_i / \partial x_i$, it duplicates the existing $\partial u_i / \partial x_i$.

(b) Equivalent candidates due to commutative property (e.g., $u_i u_j \partial u_k / \partial x_k$ is equivalent to $u_j u_i \partial u_k / \partial x_k$) or symmetry (e.g., $\partial^2 u_i / (\partial x_i \partial x_j)$ is equivalent to $\partial^2 u_i / (\partial x_j \partial x_i)$).

Through these processing steps above, the algorithm is ensured to automatically generate candidates that satisfy both validity and uniqueness.

### v. Contraction

For each candidate derived from step iv, compute its value by applying contraction operations.

### vi. Candidate library generation

Stack the values of each candidate across different dimensions row-wise and concatenate the candidates column-wise to form the candidate library matrix $\mathbf{\Theta}$.

It is noteworthy that the proposed candidate library construction method relies solely on limited a priori knowledge during the first step. No manual intervention is required in subsequent steps. Moreover, although the construction process involves numerous combinations and filtration of expressions, the primary steps (i~iv) are performed symbolically rather than numerically, ensuring high efficiency even when processing a large number of expressions.

Figure 2(b) illustrates the complete process—from construction to discovery—of the convection term in the Navier–Stokes equation. First, a convection term template is generated by combining the velocity with its first-order derivatives, yielding $3^3$ = 27 suffix assignment schemes (see Table A in Fig. 2(b)). The suffixes of these candidates are then reordered (see Table B in Fig. 2(b)), where reordered suffixes are highlighted in blue. Invalid candidates are marked with red diagonal lines, and duplicate and equivalent candidates are marked with purple diagonal lines; after these eliminations, only 3 candidates remain (see Table C in Fig. 2(b)). Finally, contraction operations are applied to compute the candidate values, forming the matrix $\mathbf{\Theta}$, and sparse regression is used to derive the convection term.

As shown in Fig. 2(b), while a conventional scalar-based approach would generate 27 candidates corresponding to the 27 suffix assignment schemes, the proposed CTSR method yields far fewer candidates. Moreover, with $n_s$ sampling points, the CTSR method produces first-order tensor candidates with 3 components that are stacked row-wise to form a candidate matrix with $3n_s$ rows. In contrast, the scalar-based method generates three separate candidate matrices for every component equation, each with $n_s$ rows. Consequently, the CTSR method improves data utilization efficiency by a factor of 3 compared to the scalar-based approach.

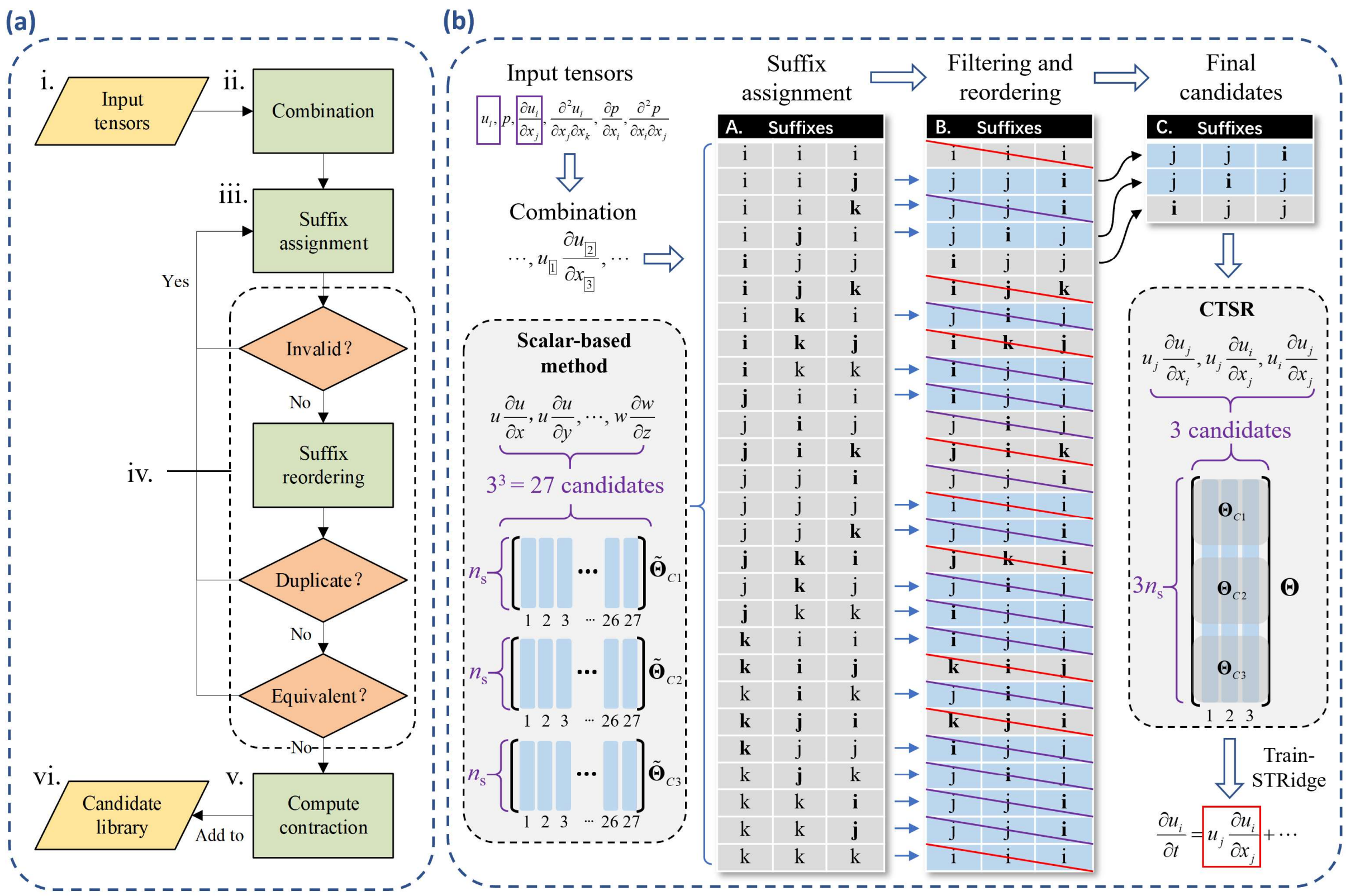

**Fig. 2.** Construction of the candidate library. (a) Flowchart; (b) Complete process from constructing the convection term to its discovery.

## 3. Sparse regression

After constructing the matrix $\mathbf{\Theta}$, the next step is to determine the coefficients of the candidate terms, denoted by $\xi$. Since $\xi$ is typically sparse—with only a few nonzero entries—a standard least-squares approach is ineffective because it tends to produce dense solutions, resulting in overly complex equations.

Sparse regression effectively addresses this issue [22,34,35,39]. Rudy et al. [22] introduced the sequence threshold ridge regression (STRidge) to solve sparse regression problems. In STRidge, a least-squares optimization augmented with an $l^2$ regularization term (i.e., ridge regression) is used to obtain an initial solution, and sparsity is enforced by applying hard-thresholding to $\xi$. The ridge regression is formulated as

$$\hat{\xi} = \arg\min_{\xi} \left\| \mathbf{\Theta}\xi - \mathbf{u}_t \right\|_2^2 + \lambda \left\| \xi \right\|_2 \tag{10}$$

where $\lambda$ represents the regularization coefficient that controls the sparsity of the solution, and $\| \cdot \|_2$ denotes the $l^2$ norm.

In STRidge, the hard-thresholding tolerance determines the sparsity of the final solution. To automatically find the optimal tolerance, Rudy also proposed a training-based extension called TrainSTRidge, which has been applied in several studies [27,40,41]. An overview of the TrainSTRidge implementation is as follows:

(a) Given the matrix $\mathbf{\Theta}$, regularization coefficient $\lambda$, and an initial tolerance $d_{\text{tol}}$, randomly split $\mathbf{\Theta}$ into training and testing sets in a 4:1 ratio.

(b) Perform a least-squares regression on the training set to obtain an initial estimate $\hat{\xi}_0$, and compute the corresponding error $E_1$ on the test set. The error is defined as

$$E = \left\| \mathbf{\Theta}^{\text{test}} \hat{\xi} - \mathbf{u}_t^{\text{test}} \right\|_2^2 + 10^{-3} \kappa(\mathbf{\Theta}) \left\| \hat{\xi} \right\|_0 \tag{11}$$

where $\kappa(\mathbf{\Theta})$ denotes the condition number of $\mathbf{\Theta}$, and $\| \cdot \|_0$ denotes the $l^0$ norm, which counts the number of nonzero elements.

(c) On the training set, perform STRidge: first, obtain $\hat{\xi}$ via ridge regression as in Eq. (10); then, apply hard-threshold to discard elements of $\xi$ below the tolerance. This procedure is repeated until no remaining elements exceed the tolerance or the maximum number of iterations, $n_{\text{STRidge}}$, is reached. Next, compute the testing error $E_2$ as defined in Eq. (11).

(d) If $E_2$ is lower than $E_1$ (indicating improvement), decrease the tolerance and update $E_1$ to $E_2$; if $E_2$ is higher, increase the tolerance. Repeat steps (c) ~ (d) until reaching the maximum number of training steps, $n_{\text{train}}$, and then output the final solution $\hat{\xi}$.

In this study, TrainSTRidge is used to determine the sparse vector $\hat{\xi}$. Further details about the algorithm can be found in Ref. [22].

## C. Hyperparameter setup

The performance of the proposed method is influenced by the selection of hyperparameters. The CTSR method involves four primary hyperparameters: the regularization coefficient $\lambda$, the maximum number of training steps $n_{\text{train}}$, the number of STRidge iterations $n_{\text{STRidge}}$, and the initial tolerance $d_{\text{tol}}$. Our experiments indicate that $\lambda$, $n_{\text{train}}$, and $n_{\text{STRidge}}$ have a minor impact on the results. In this study, $\lambda$ is set to $1\times10^{-5}$, while $n_{\text{train}}$ and $n_{\text{STRidge}}$ are set to 25 and 10, respectively.

In contrast, the selection of initial tolerance $d_{\text{tol}}$ is critical for accurately discovering the governing equations. A higher $d_{\text{tol}}$ typically produces a sparser $\hat{\xi}$ and results in simpler equations, whereas a lower $d_{\text{tol}}$ generally leads to equations with a larger number of terms. Here, $d_{\text{tol}}$ is determined via Pareto front analysis to optimally balance accuracy and conciseness (see Appendix A for details). Specifically, $d_{\text{tol}}$ is set to 0.001, 0.01, 0.01, and 1.2 for the four test cases presented in the next section, respectively.

Table I summarizes the hyperparameter values used in this study.

**Table I.** Hyperparameter selection.

| Hyperparameter | Value |
|---|---|
| $\lambda$ | $1\times10^{-5}$ |
| $d_{\text{tol}}$ | 0.001、0.01、0.01、1.2 |
| $n_{\text{train}}$ | 25 |
| $n_{\text{STRidge}}$ | 10 |

# III. Results

To validate the accuracy of the proposed method, we applied it to discover four governing equations with increasing complexity and dimensionality. These include: (1) the 2D Burgers equation governing random vortex evolution; (2) the 2D Navier–Stokes equation with a source term for natural convection in a square cavity; (3) the 3D Navier–Stokes equation for flow over a cylinder; and (4) the 3D Giesekus equation modeling viscoelastic blood flow in a realistic cerebral artery model. The first three cases involve first-order equations with two components in 2D or three components in 3D, whereas the fourth case is a second-order equation with nine components. Table II summarizes the basic information for each case.

**Table II.** Overview of the test cases.

| | Governing equation | Problem | Dimension | Order |
|---|---|---|---|---|
| Case 1 | Burgers equation | Random vortex evolution | 2D | 1 |
| Case 2 | Navier-Stokes equation with source term | Natural convection in a square cavity | 2D | 1 |
| Case 3 | Navier-Stokes equation | Flow over a cylinder | 3D | 1 |
| Case 4 | Giesekus equation | Viscoelastic blood flow in a realistic cerebral artery model | 3D | 2 |

## A. 2D Burgers equation for random vortex evolution

The Burgers equation plays an important role in fluid mechanics and acoustics and is widely used as a benchmark for data-driven discovery of governing equations. With a structure closely resembling that of the Navier-Stokes equations, and it encapsulates two fundamental transport mechanisms in fluid dynamics: nonlinear convection and viscous diffusion. As such, reliably identifying the Burgers equation is considered an important prerequisite for tackling more complex problems in fluid dynamics. Given that most existing data-driven methods have accurately discovered the one-dimensional (1D) Burgers equation [19,42,43], we employ the more complex 2D Burgers equation as the first test case. The added spatial dimension allows for the description of more intricate and realistic flow phenomena, such as vortex dynamics.

In Cartesian tensor notation, the 2D Burgers equation is expressed as:

$$\frac{\partial u_i}{\partial t} = -u_j \frac{\partial u_i}{\partial x_j} + \varepsilon \frac{\partial^2 u_i}{\partial x_j \partial x_j} \tag{12}$$

where $\varepsilon$ denotes the diffusion coefficient. The dataset for equation discovery is computed using the finite-difference solver Incompact3d [44]. The simulation is conducted in a computational domain discretized on a uniform 128×128 grid with periodic boundary conditions. The convection and diffusion terms are discretized using a second-order central difference scheme, while time integration is performed with the Euler method.

The dataset consists of 200 time-steps, each spaced by $\Delta t = 0.02$, with $\varepsilon$ set to 0.1. The initial condition is defined as [15]:

$$\omega(x,y)=\frac{2\omega_0(x,y)}{\max|\omega_0|}+c$$
$$\omega_0(x,y)=\sum_{|k|,|l|\le 4}\lambda_{k,l}\cos(kx+ly)+\gamma_{k,l}\sin(kx+ly) \tag{13}$$

where $\omega$ denotes the vorticity; $\lambda_{k,l}$ and $\gamma_{k,l}$ are independent random variables drawn from a normal distribution (i.e., $\lambda_{k,l},\gamma_{k,l}\sim\mathcal{N}(0,1)$), and $c$ is uniformly distributed over [–2, 2] (i.e., $c\sim\mathcal{U}(-2,2)$). Figure 3 shows the velocity field distributions at various time instants obtained from the numerical simulation.

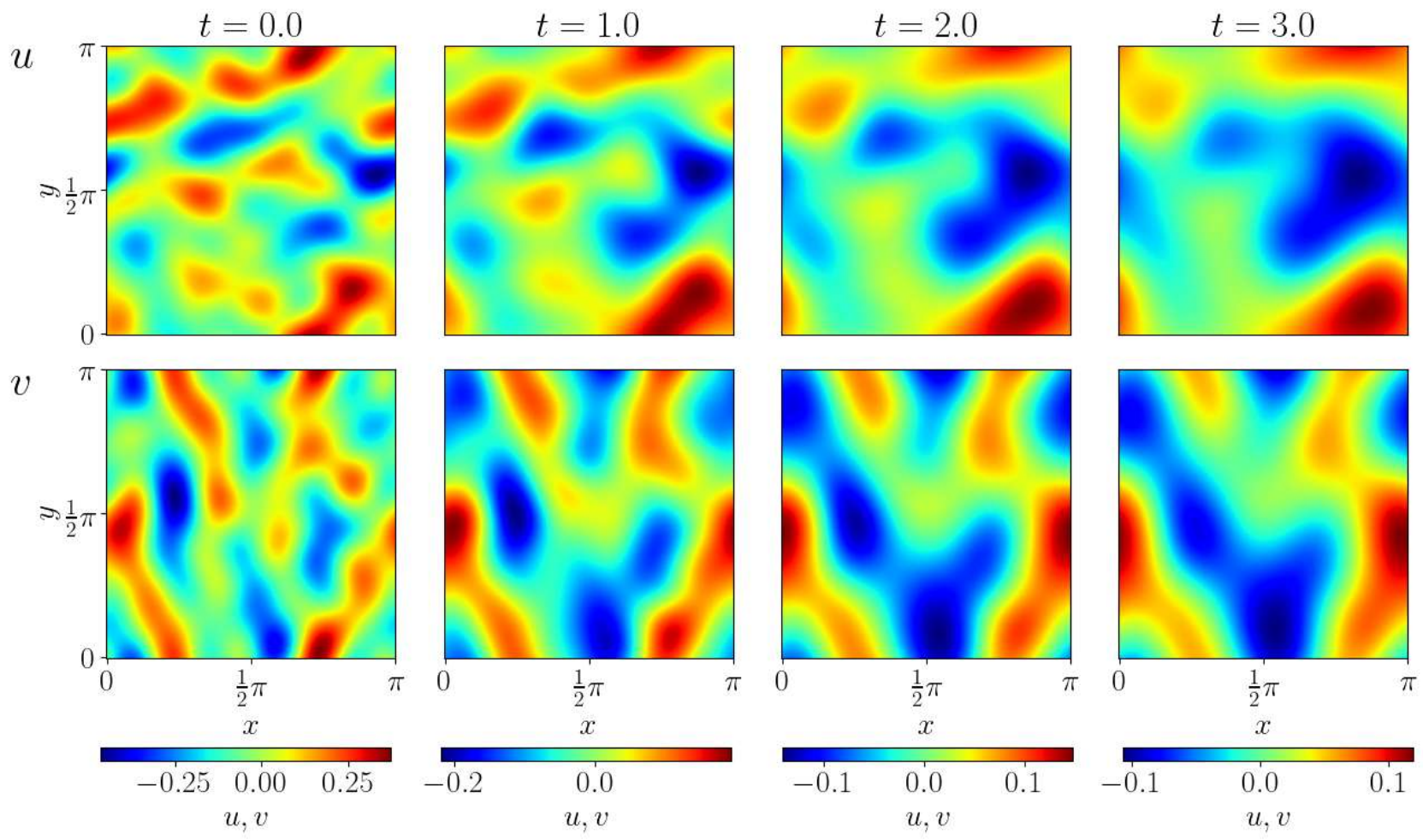

**Fig. 3.** Velocity distribution for the 2D Burgers case.

To evaluate the effectiveness of the CTSR algorithm under sparse data conditions, the dataset for equation discovery is generated by randomly subsampling spatio-temporal data from numerical results. Specifically, 50 spatial points and 20 temporal points are sampled. In constructing the candidate library, the velocity along with its first- and second-order derivatives are selected as input tensors. Derivatives are computed using second-order central difference scheme for interior nodes and second-order one-sided difference scheme for boundary nodes. To minimize the influence of boundary artifacts, data at spatial boundary nodes are excluded from the sampling process. The parameter $P$ is set to 2 to introduce sufficient complexity, yielding 13 candidates:

$$\left\{\frac{\partial^2 u_j}{\partial x_i\partial x_j},\ \frac{\partial^2 u_i}{\partial x_j\partial x_j},u_i,\cdots,u_ju_k\frac{\partial^2 u_i}{\partial x_j\partial x_k},u_ju_i\frac{\partial^2 u_k}{\partial x_j\partial x_k}\right\} \tag{14}$$

For comparison, we employed the well-known sparse regression method PDE-FIND [22] as a benchmark, which has been widely used as a baseline in previous studies [19,32,45]. Since PDE-FIND shares the same hyperparameters and sparse regression framework as CTSR, it enables a fair and transparent comparison while highlighting the influence of the Cartesian tensor-based strategy on equation discovery. Operating on a scalar basis, PDE-FIND generates 77 candidates for $P$ = 2:

$$\left\{u, v, u^2, uv, v^2, \cdots, v^2\frac{\partial^2 v}{\partial y \partial x}, v^2\frac{\partial^2 v}{\partial y \partial y}\right\} \tag{15}$$

To quantify accuracy, we define the prediction error as the mean relative error between the identified coefficients for the terms in the exact governing equation and their true values. Additionally, redundancy is quantified by counting the nonzero terms in the discovered equation that do not appear in the exact formulation.

The hyperparameters for CTSR are listed in Table III, and its initial tolerance $d_{\text{tol}}$ is determined to be 0.001 via Pareto front analysis (see Appendix A). For a fair and objective comparison, the values of $d_{\text{tol}}$ for PDE-FIND are selected by minimizing the prediction error (assuming the exact form of the equation is known). Specifically, if PDE-FIND successfully identifies the correct equation structure, $d_{\text{tol}}$ is chosen to minimize the error for that structure; if not, $d_{\text{tol}}$ is selected from the set of tested values based on the minimum error. All other hyperparameters for PDE-FIND are set to be consistent with those used in CTSR.

Table III summarizes the discovered results for the 2D Burgers equation obtained using both methods, reporting the prediction error and the number of redundant terms. Since PDE-FIND operates on a scalar basis, the regression is performed separately for the $x$- and $y$-component equations; separate results are provided for each component. As shown, both CTSR and PDE-FIND accurately discover the structure of the equation without introducing redundant terms. In particular, the CTSR method achieves a prediction error of 0.014%, which is lower than the errors of 0.041% and 0.151% obtained for the corresponding components by PDE-FIND. These results demonstrate that, for this relatively simple test case, both approaches successfully discover the 2D Burgers equation, with CTSR exhibiting slightly higher accuracy.

**Table III.** Summary of discovered results for the 2D Burgers equation.

| **Exact form** | | $\frac{\partial u_i}{\partial t} = -u_j\frac{\partial u_i}{\partial x_j} + 0.1\frac{\partial^2 u_i}{\partial x_j \partial x_j}$ | | |
|---|---|---|---|---|
| **Method** | **No. of candidates** | **Discovered equation** | **Error** | **No. of redundant terms** |
| CTSR | 13 | $\frac{\partial u_i}{\partial t} = -0.99982u_j\frac{\partial u_i}{\partial x_j} + 0.10002\frac{\partial^2 u_i}{\partial x_j \partial x_j}$ | 0.014% | 0 |
| PDE-FIND | 77 | $\frac{\partial u}{\partial t} = -1.0006u\frac{\partial u}{\partial x} - 1.0003v\frac{\partial u}{\partial y} + 0.10007\frac{\partial^2 u}{\partial x \partial x} + 0.099997\frac{\partial^2 u}{\partial y \partial y}$ | 0.041% | 0 |
| | | $\frac{\partial v}{\partial t} = -0.99927u\frac{\partial v}{\partial x} - 0.99702v\frac{\partial v}{\partial y} + 0.10003\frac{\partial^2 v}{\partial x \partial x} + 0.099798\frac{\partial^2 v}{\partial y \partial y}$ | 0.151% | 0 |

## B. 2D Navier–Stokes equation for natural convection

The second test case examines a more complex flow scenario—natural convection in a 2D square cavity, as schematically illustrated in Fig. 4. In this configuration, the left wall of the cavity is heated while the right wall is cooled, thereby inducing natural convection through buoyancy effects. The governing equation is the 2D Navier–Stokes equation with an added buoyancy source term, which is expressed in Cartesian tensor notation as [46]

$$\frac{\partial u_i}{\partial t} = -u_j \frac{\partial u_i}{\partial x_j} + \mathrm{Pr}\,\mathrm{Ra}^{-1/2} \frac{\partial^2 u_i}{\partial x_j \partial x_j} - \frac{\partial p}{\partial x_i} - \mathrm{Pr}\,\theta g_i \tag{16}$$

where $g_i$, $p$, and $\theta$ denote the normalized gravitational acceleration (i.e., $g_i = [g_x, g_y] = [0, -1]$), pressure, and temperature, respectively, while Pr and Ra represent the Prandtl and Rayleigh numbers.

The Navier–Stokes equation forms the foundation of fluid mechanics. By incorporating appropriate source terms to account for external forces or additional physical processes, it can describe a wide range of complex flow phenomena such as atmospheric circulation and plasma dynamics. As such, this test case offers a representative benchmark to assess CTSR's potential for discovering governing equations in fluid systems with coupled physical effects.

The dataset is computed numerically using Incompact3d over a domain defined by $x, y \in [0, 1]$. The left and right boundaries are set to hot and cold wall conditions, respectively, and the top and bottom boundaries are imposed with adiabatic conditions. The simulation is conducted at Ra = $10^6$ and Pr = 0.71. Initially, the temperature field is prescribed as a linear function along the $x$-direction to satisfy $\theta(x) = 0.5 - x$. The governing equation is numerically solved on a uniform 193×193 grid using a fourth-order compact finite difference scheme for spatial discretization and a third-order Adams–Bashforth scheme for time integration. The dataset comprises 300 time-steps with a time interval of $\Delta t = 0.03$. Figure 5 shows the temperature and velocity fields, where the velocity vectors are indicated by arrows.

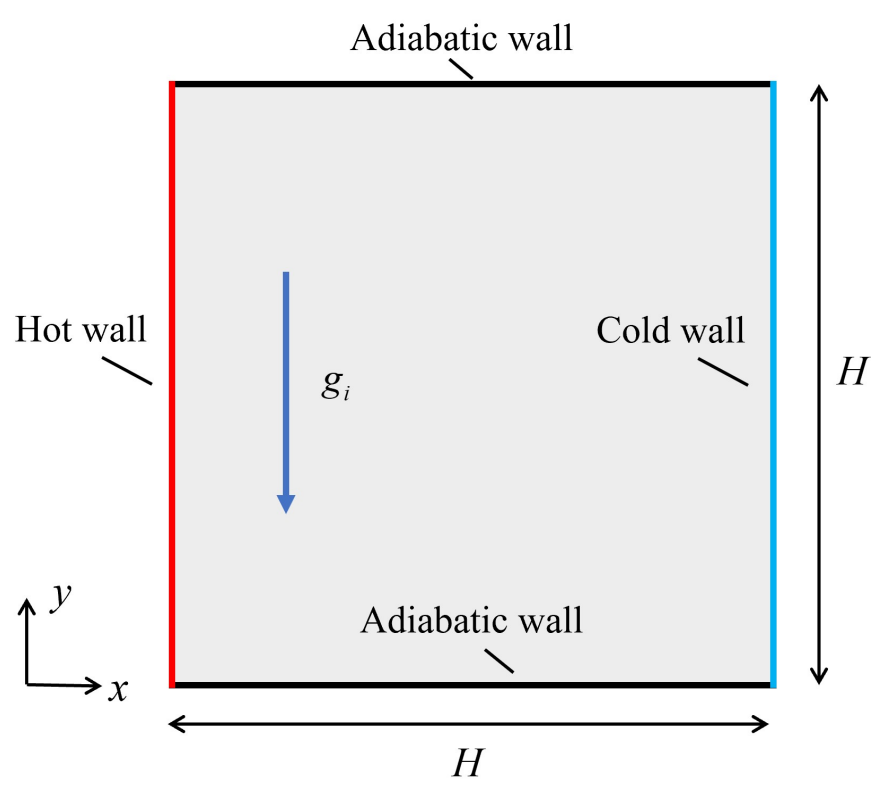

**Fig. 4.** Schematic representation of natural convection in a square cavity.

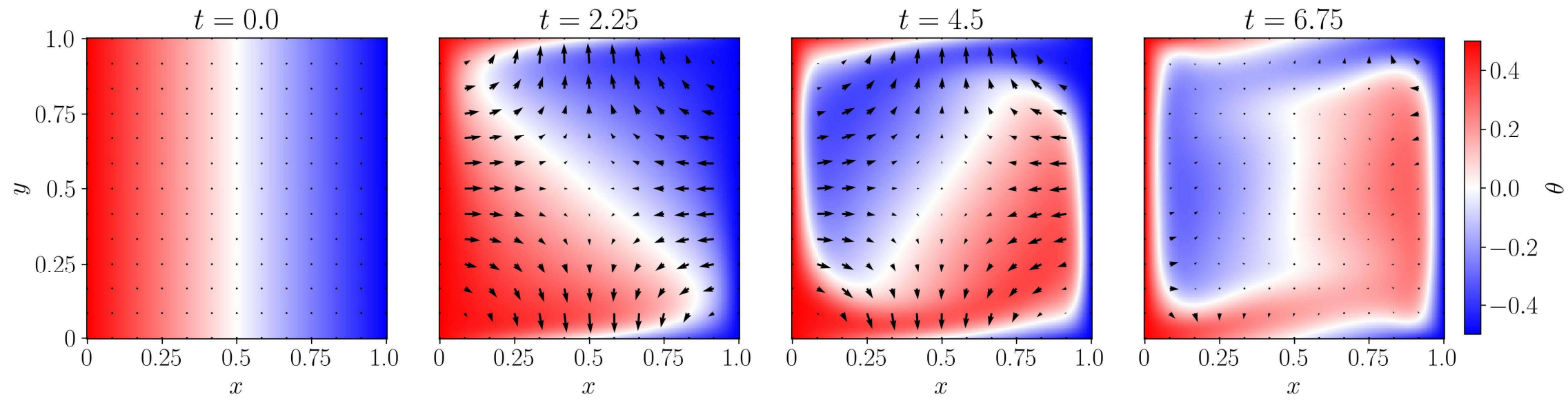

**Fig. 5.** Temperature and velocity distribution in the cavity.

The dataset used for equation discovery is randomly sampled from both the spatial and temporal domains, with 50 samples in space and 20 in time. In constructing the candidate library, the velocity, pressure, and temperature—along with their first and second spatial derivatives are used as input tensors. The computation of derivatives and the treatment of boundary node data remain consistent with Case 1. Additionally, $g_i$ is also included among the inputs for CTSR. With $P = 2$, a total of 70 candidates are generated:

$$\left\{ \frac{\partial^2 u_j}{\partial x_i \partial x_j}, \frac{\partial^2 u_i}{\partial x_j \partial x_j}, \frac{\partial p}{\partial x_i}, \cdots, \theta^2 \frac{\partial \theta}{\partial x_i}, \theta^2 g_i \right\} \tag{17}$$

The PDE-FIND method produces 374 candidate functions when $P = 2$, approximately 5 times as many as the CTSR. The initial tolerance $d_{\mathrm{tol}}$ for CTSR is set to 0.01, while all other hyperparameters remained unchanged.

Compared to Case 1, this test case, while still 2D, involves a more complex governing equation with additional variables and exhibits less spatial richness, thereby imposing more stringent requirements on the discovery method. Table IV summarizes the discovered results for the 2D Navier–Stokes equation obtained via CTSR and PDE-FIND. The CTSR method accurately discovered the equation form and precisely identified the coefficients, achieving a prediction error of 0.64%. In contrast, the PDE-FIND method failed to accurately discover the equation form. Specifically, for the $x$-component, PDE-FIND discovered all the correct terms but also generated 15 redundant terms; for the $y$-component, it erroneously omitted the diffusion term along the $y$-axis while generating one redundant term. The corresponding prediction errors were 9.74% and 34.8%, respectively, substantially exceeding those achieved by CTSR.

It is noteworthy that despite the correct identification of all terms in the $x$-component equation by PDE-FIND, a notable number of redundant terms complicate the extraction of meaningful information, particularly when the exact form of the governing equation is unknown.

**Table IV.** Summary of discovered results for the 2D Navier-Stokes equation.

| **Exact form** | | $\frac{\partial u_i}{\partial t} = -u_j \frac{\partial u_i}{\partial x_j} + 0.00071 \frac{\partial^2 u_i}{\partial x_j \partial x_j} - \frac{\partial p}{\partial x_i} + 0.71\theta g_i$ | | |
|---|---|---|---|---|
| **Method** | **No. of candidates** | **Discovered equation** | **Error** | **No. of redundant terms** |
| CTSR | 70 | $\frac{\partial u_i}{\partial t} = -1.00 u_j \frac{\partial u_i}{\partial x_j} + 0.000698 \frac{\partial^2 u_i}{\partial x_j \partial x_j} - 0.997 \frac{\partial p}{\partial x_i} + 0.706\theta g_i$ | 0.64% | 0 |
| PDE-FIND | 374 | $\frac{\partial u}{\partial t} = -0.558 u \frac{\partial u}{\partial x} - 0.995 v \frac{\partial u}{\partial y} + 0.000692 \frac{\partial^2 u}{\partial x \partial x} + 0.000703 \frac{\partial^2 u}{\partial y \partial y} - 0.995 \frac{\partial p}{\partial x} +$ | 9.74% | 15 |
| | | $\frac{\partial v}{\partial t} = -0.971 u \frac{\partial v}{\partial x} - 0 v \frac{\partial v}{\partial y} + 0.000725 \frac{\partial^2 v}{\partial x \partial x} + 0 \frac{\partial^2 v}{\partial y \partial y} - 0.978 \frac{\partial p}{\partial y} + 0.699\theta g_y$ | 34.8% | 1 |

## C. 3D Navier-Stokes equation for flow over a cylinder

Previous studies on data-driven equation discovery have frequently employed 2D flow over a cylinder as a test case [10,21,22]. Typically, these studies focus on learning the 2D vorticity transport equation, which neglects the vortex stretching term—a key mechanism for turbulence generation. Therefore, discovering the complete 3D vorticity or Navier–Stokes equation from data is critical for advancing turbulence research using data-driven methods. Moreover, as most real-world fluid flows in natural and engineering applications are inherently three-dimensional, directly uncovering the 3D form of the Navier–Stokes equation offers greater practical significance.

In the present work, the 3D Navier–Stokes equation is discovered from cylinder wake data. The nondimensional form of the equation is given by,

$$\frac{\partial u_i}{\partial t} = -u_j \frac{\partial u_i}{\partial x_j} + \frac{1}{\mathrm{Re}} \frac{\partial^2 u_i}{\partial x_j \partial x_j} - \frac{\partial p}{\partial x_i} \tag{2}$$

Here, Re is based on the cylinder diameter $D$ and is set to 200. At this Reynolds number, the results reported in Ref. [47] demonstrate the presence of a distinct spanwise flow in the cylinder wake, which confirms the emergence of 3D flow structures.

Figure 6 illustrates the schematic of this case. The flow enters along the positive $x$-axis, with the cylinder's axis aligned with the $z$-axis. The computational domain measures $20D \times 12D \times 6D$, discretized using a $385 \times 192 \times 32$ uniform grid. Dirichlet boundary conditions are applied at the inlet and outlet, whereas periodic boundary conditions are applied to the remaining external boundaries. The cylinder surface is treated as a no-slip wall via an immersed boundary method. The governing equation is solved using the Incompact3d solver, which employs a sixth-order compact scheme for spatial discretization and a third-order Adams–Bashforth scheme for time integration.

The dataset for governing equation discovery is generated by random subsampling within the cylinder wake region. The subsampling domain measures approximately $6.7D \times 4D \times 6D$, with its left boundary located $6.65D$ from the inlet and its top and bottom boundaries situated $4D$ from the corresponding external boundaries. A total of 50 spatial and 20 temporal sampling points are used. The velocity distribution obtained from the simulation is illustrated in Fig. 7.

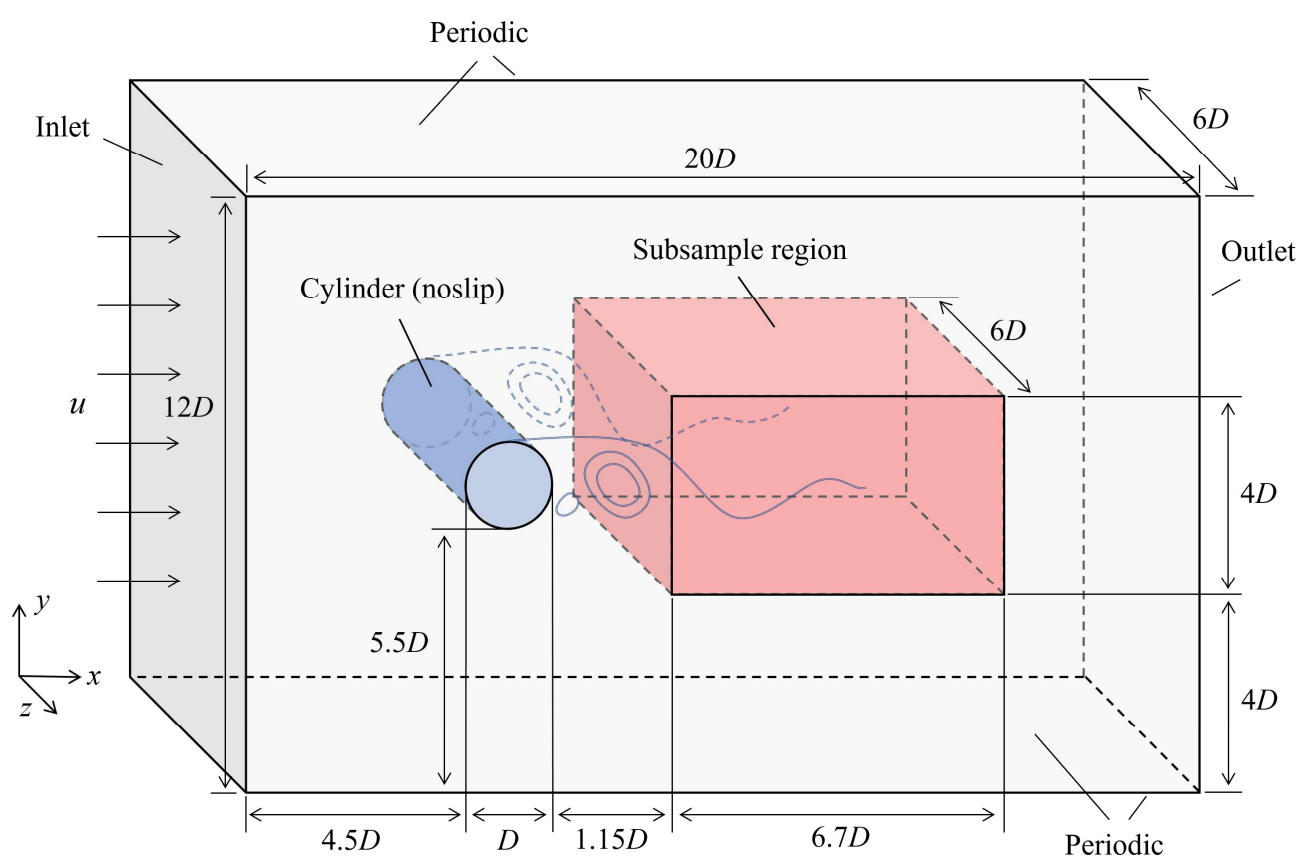

**Fig. 6.** Schematic illustration of flow around a cylinder (not drawn to scale).

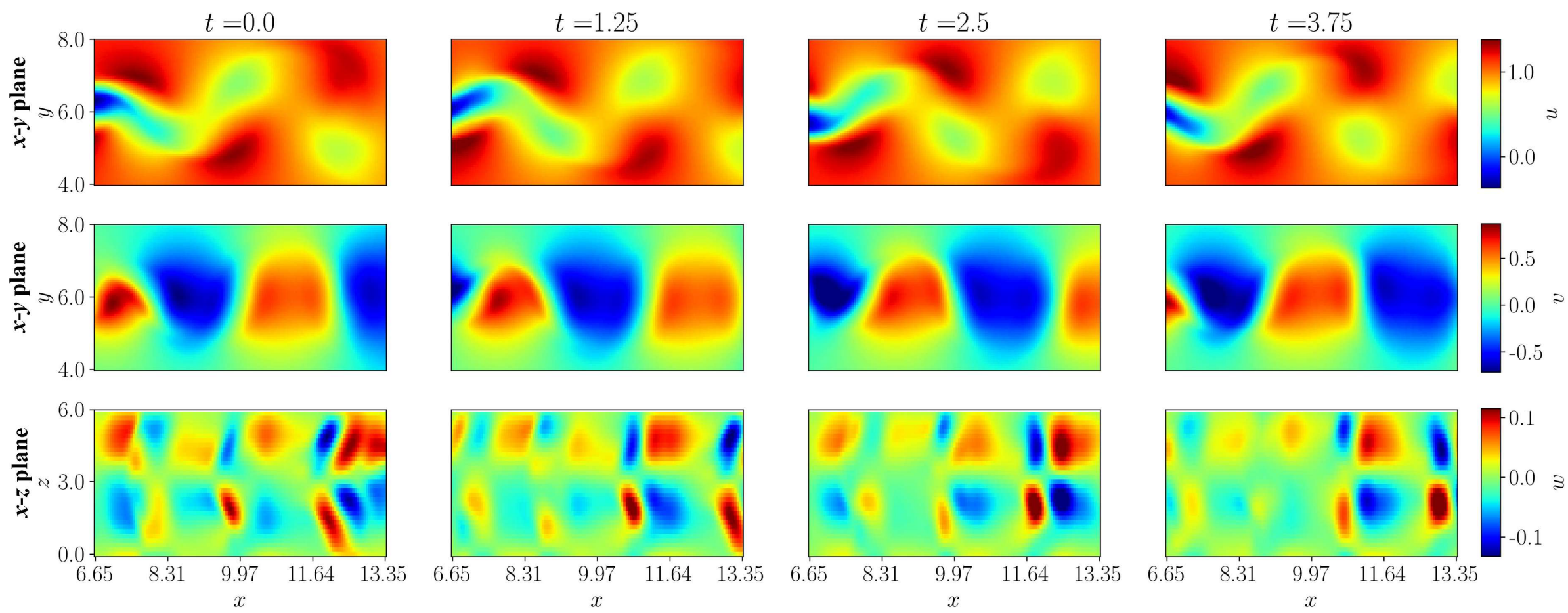

**Fig. 7.** Velocity distribution for the cylinder wake flow.

In this case, the input tensor comprises the velocity, pressure, and their first- and second-order derivatives. The computation of derivatives and the treatment of boundary node data remain consistent with Case 1. With $P = 2$, a total of 30 candidate terms are generated, which are

$$\left\{ \frac{\partial^2 u_j}{\partial x_i \partial x_j}, \frac{\partial^2 u_i}{\partial x_j \partial x_j}, \frac{\partial p}{\partial x_i}, \cdots, p^2 \frac{\partial^2 u_i}{\partial x_j \partial x_j}, p^2 \frac{\partial p}{\partial x_i} \right\} \tag{18}$$

While the PDE-FIND method generates 734 candidate functions for $P = 2$, approximately 20 times more than those generated by CTSR. For CTSR, the initial tolerance, $d_{\text{tol}}$, is set to 0.01, while the other hyperparameters remain unchanged.

Table V summarizes the discovered results for the 3D Navier–Stokes equation obtained via CTSR and PDE-FIND. The CTSR method accurately identified both the structure of the equations and the associated coefficients, achieving a prediction error of 0.57%. This result demonstrates that CTSR maintains a high level of accuracy even in a relatively complex 3D problem. In contrast, PDE-FIND failed to capture the correct structure of the component equations in all three spatial directions, as evidenced by the prediction of multiple redundant terms. Additionally, PDE-FIND incurred significantly higher prediction errors, ranging from 20.2% to 32.8%.

**Table V.** Summary of discovered results for the 3D Navier-Stokes equation.

| **Exact form** | | $\frac{\partial u_i}{\partial t} = -u_j \frac{\partial u_i}{\partial x_j} + 0.005 \frac{\partial^2 u_i}{\partial x_j \partial x_j} - \frac{\partial p}{\partial x_i}$ | | |
|---|---|---|---|---|
| **Method** | **No. of candidates** | **Discovered equation** | **Error** | **No. of redundant terms** |
| CTSR | 30 | $\frac{\partial u_i}{\partial t} = -0.998 u_j \frac{\partial u_i}{\partial x_j} + 0.00493 \frac{\partial^2 u_i}{\partial x_j \partial x_j} - 0.999 \frac{\partial p}{\partial x_i}$ | 0.57% | 0 |
| PDE-FIND | 734 | $\frac{\partial u}{\partial t} = -0.828 u \frac{\partial u}{\partial x} - 1.00 v \frac{\partial u}{\partial y} - 0 w \frac{\partial u}{\partial z} + 0.00514 \frac{\partial^2 u}{\partial x \partial x} + 0.00480 \frac{\partial^2 u}{\partial y \partial y} + 0 \frac{\partial^2 u}{\partial z \partial z} - 0.999 \frac{\partial p}{\partial x} + \cdots$ | 32.0% | 6 |
| | | $\frac{\partial v}{\partial t} = -0.998 u \frac{\partial v}{\partial x} - 0.779 v \frac{\partial v}{\partial y} - 1.08 w \frac{\partial v}{\partial z} + 0.00513 \frac{\partial^2 v}{\partial x \partial x} + 0.00457 \frac{\partial^2 v}{\partial y \partial y} + 0 \frac{\partial^2 v}{\partial z \partial z} - 1.00 \frac{\partial p}{\partial y} + \cdots$ | 20.2% | 8 |
| | | $\frac{\partial w}{\partial t} = -1.00 u \frac{\partial w}{\partial x} - 0.951 v \frac{\partial w}{\partial y} - 0 w \frac{\partial w}{\partial z} + 0.00530 \frac{\partial^2 w}{\partial x \partial x} + 0.00584 \frac{\partial^2 w}{\partial y \partial y} + 0 \frac{\partial^2 w}{\partial z \partial z} - 1.02 \frac{\partial p}{\partial z} + \cdots$ | 32.8% | 4 |

Notably, the discovered equation for the $z$-direction by PDE-FIND exhibited the lowest accuracy, as evidenced by both a higher prediction error and a larger number of redundant terms. Furthermore, PDE-FIND nearly omitted all terms involving $z$-directional derivatives, including terms in the convective (e.g., $w\partial u/\partial z$ and $w\partial w/\partial z$ ) and diffusive (e.g., $\partial^2 u/(\partial z\partial z)$ , $\partial^2 v/(\partial z\partial z)$ and $\partial^2 w/(\partial z\partial z)$ ). This phenomenon can be attributed to the relatively low Reynolds number; under this condition, the streamwise and vertical flows dominate, resulting in a sparse distribution of flow features along the $z$-direction. This sparsity hinders the accurate discovery of the $z$-direction equation and its associated terms. Further discussion on this issue is provided in Section IV.

### D. 3D Giesekus equation for viscoelastic blood flow

Finally, we consider a more complex case—the 3D viscoelastic blood flow in a realistic cerebral artery model. Blood is a complex mixture comprising proteins, red blood cells, platelets, white blood cells, and other components. In the human microcirculatory system, blood exhibits significant non-Newtonian or viscoelastic properties [48]. Viscoelastic fluids combine the characteristics of elastic solids and viscous fluids, exhibiting behaviors such as shear thinning. The total stress, $\tau_{ij}$, experienced by a viscoelastic fluid can be decomposed into a solvent contribution, $\tau_{ij}^s$, and a polymer contribution, $\tau_{ij}^p$. Typically, a linear relationship is assumed between $\tau_{ij}^s$ and the shear rate $\dot{\gamma}_{ij}$, whereas $\tau_{ij}^p$ can be obtained from viscoelastic models. Representative models include the Oldroyd-B model [49], the Giesekus model [50], and the Phan-Thien–Tanner (PTT) model [51]. Previous studies have demonstrated the effectiveness of the Giesekus model in predicting the rheological behavior of blood [48]. In this case, we employ the Giesekus model to compute $\tau_{ij}^p$, and treat it as the target equation for discovery. Under steady-state conditions, the Giesekus equation is expressed as follows [50]:

$$\eta_p \dot{\gamma}_{ij} = \tau_{ij}^p + \lambda_1 \left( u_k \frac{\partial \tau_{ij}^p}{\partial x_k} - \tau_{ik}^p \frac{\partial u_j}{\partial x_k} - \frac{\partial u_i}{\partial x_k} \tau_{kj}^p \right) + \alpha \frac{\lambda_1}{\eta_p} \tau_{ik}^p \tau_{kj}^p \tag{19}$$

where $\eta_p$ denotes the polymer contribution to the viscosity, $\lambda_1$ is the relaxation time, $\alpha$ is the mobility factor, and the shear rate tensor $\dot{\gamma}_{ij}$ is defined as:

$$\dot{\gamma}_{ij} = \frac{\partial u_i}{\partial x_j} + \frac{\partial u_j}{\partial x_i} \tag{20}$$

As a nonlinear second-order tensor equation, the Giesekus model provides sufficient mathematical complexity to serve as a stringent benchmark for evaluating CTSR's capability to discover complex governing equations. More importantly, it belongs to a critical class of equations in fluid mechanics—constitutive equations—that define the relationship between stress and deformation. Constitutive discovery and modeling remain significant challenges in areas such as viscoelastic flows and granular media. This case serves as a preliminary exploration of CTSR's ability to identify complex constitutive relationships, which may potentially contribute to discovering unknown constitutive relations or enhancing existing constitutive models in future applications.

The dataset is obtained from numerical simulations using rheoFoam [52], a viscoelastic fluid solver built on OpenFOAM®. According to Ref. [53], the blood parameters in Eq. (19) are set as follows: $\eta_p$ = 0.0043 Pa·s, $\lambda_1$ = 0.008 s, and $\alpha$ = 0.5. The arterial model (AHMU1218100) used in this case is derived from computed tomography angiography (CTA) as provided in Ref. [54]. To minimize the effects of inlet/outlet disturbances on arterial flow, circular tubes are extended at both the inlet and outlet of the artery. The inlet radius is $R_{\text{in}}$ = 1.042 mm, while the outlet radii are $R_{\text{out1}}$ = 0.656 mm and $R_{\text{out2}}$ = 0.961 mm, respectively. According to Ref. [54], the average mass flow rate at the inlet is $\dot{m}_{in}$ = 1.336 $\times 10^{-3}$ kg/s. Figure 8(a) shows the arterial model, the imposed boundary conditions, and the computed streamlines, while Figure 8(b) shows the distributions of the six independent components of the symmetric second-order tensor $\tau_{ij}^{p}$.

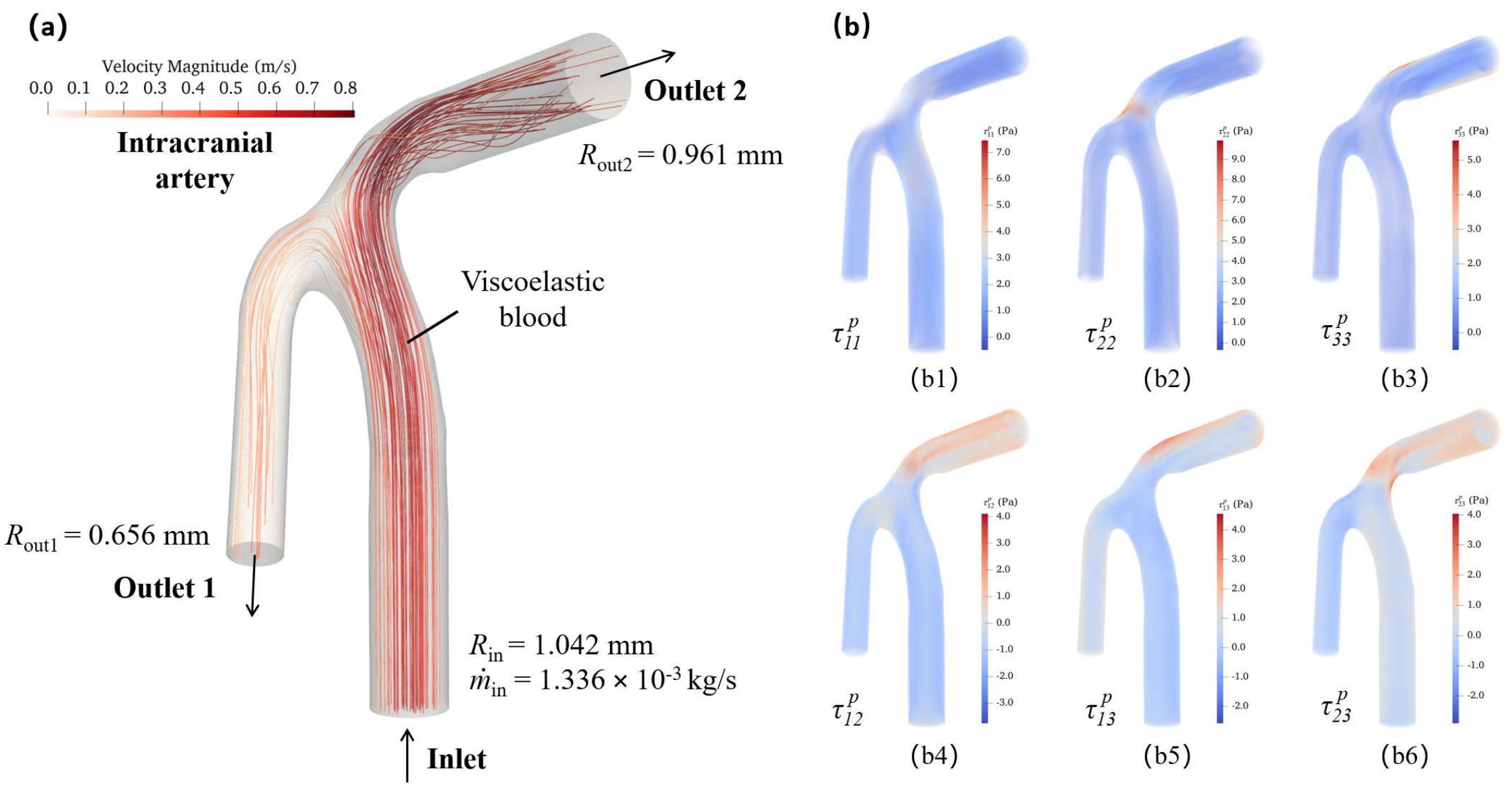

**Fig. 8.** Viscoelastic blood flow in a realistic cerebral artery model. (a) Schematic representation; (b) Distribution of $\tau_{ij}^{p}$.

In this case, the input tensor comprises the velocity, the polymer stress, and their first-order derivatives. The derivatives are computed on unstructured mesh cells using the Visualization Toolkit (VTK) [55]. To minimize the influence of derivative errors near the spatial boundaries, cells located within 0.3 mm of the boundaries are excluded during the sampling process. The dataset used for equation discovery is obtained by randomly sampling 1000 points from the space. The initial tolerance for CTSR is set to $d_{\text{tol}}$ = 1.2, and $P$ is set to 2. Since the governing equation is steady, the left-hand side term, $\eta_p \dot{\gamma}_{ij}$ is treated as $\mathbf{u}_t$ in Eq. (8), and CTSR is applied to discover the right-hand side terms in Eq. (19). To prevent redundancy with terms already appearing on the left-hand side, first-order derivatives of velocity are excluded from the candidate libraries for both CTSR and PDE-FIND. Consequently, CTSR generates 85 candidates, given by,

$$\left\{ \tau_{ij}^{p}, u_k \frac{\partial \tau_{ik}^{p}}{\partial x_j}, u_k \frac{\partial \tau_{jk}^{p}}{\partial x_i} \cdots, \tau_{jk}^{p} \tau_{ll}^{p} \frac{\partial u_k}{\partial x_i}, \tau_{jk}^{p} \tau_{ll}^{p} \frac{\partial u_i}{\partial x_k} \right\} \tag{21}$$

In contrast, with $P$ = 2, the PDE-FIND method generates 1530 candidates, roughly 13 times more than CTSR.

Since $\dot{\gamma}_{ij}$ is a symmetric second-order tensor, Eq. (19) comprises six independent component equations; in this case, PDE-FIND attempts to learn these six equations. Table VI summarizes the discovered results for the 3D Giesekus equation obtained via CTSR and PDE-FIND. It is evident that even for such a complex equation, the CTSR method accurately discovers the equation structure with an acceptable prediction error of 2.32%. In contrast, the PDE-FIND method produces over 30 redundant terms or omits necessary terms for all component equations, resulting in prediction errors exceeding 50% and demonstrating its failure in this scenario.

From the four case studies discussed above, it is evident that for complex, high-dimensional dynamical systems, the CTSR method achieves significantly higher accuracy than the PDE-FIND method.

**Table VI.** Summary of discovered results for the 3D Giesekus equation.

| **Exact form** | | $0.0043\dot{\gamma}_{ij} = \tau_{ij}^{p} + 0.008\left( u_k \frac{\partial \tau_{ij}^{p}}{\partial x_k} - \tau_{ik}^{p} \frac{\partial u_j}{\partial x_k} - \frac{\partial u_i}{\partial x_k} \tau_{kj}^{p} \right) + 0.93\tau_{ik}^{p}\tau_{kj}^{p}$ | | |
|---|---|---|---|---|
| **Method** | **No. of candidates** | **Discovered equation** | **Error** | **No. of redundant terms** |
| CTSR | 85 | $0.0043\dot{\gamma}_{ij} = 1.04\tau_{ij}^{p} + 0.00781 u_k \frac{\partial \tau_{ij}^{p}}{\partial x_k} - 0.00789\tau_{ik}^{p} \frac{\partial u_j}{\partial x_k}$ $-0.00790 \frac{\partial u_i}{\partial x_k} \tau_{kj}^{p} + 0.906\tau_{ik}^{p}\tau_{kj}^{p}$ | 2.32% | 0 |

**Table VI.** (continued).

| **Exact form** | | $0.0043\dot{\gamma}_{ij} = \tau_{ij}^{p} + 0.008\left(u_k\frac{\partial \tau_{ij}^{p}}{\partial x_k} - \tau_{ik}^{p}\frac{\partial u_j}{\partial x_k} - \frac{\partial u_i}{\partial x_k}\tau_{kj}^{p}\right) + 0.93\tau_{ik}^{p}\tau_{kj}^{p}$ | | |
|---|---|---|---|---|
| **Method** | **No. of candidates** | **Discovered equation** | **Error** | **No. of redundant terms** |
| PDE-FIND | 1530 | $0.0043\dot{\gamma}_{11} = 0.863\tau_{11}^{p} + 0.00329u\frac{\partial \tau_{11}^{p}}{\partial x} + 0.00359v\frac{\partial \tau_{11}^{p}}{\partial y} + 0.00373w\frac{\partial \tau_{11}^{p}}{\partial z} - 0.00540\tau_{11}^{p}\frac{\partial u}{\partial x} - 0.00556\tau_{12}^{p}\frac{\partial u}{\partial y} - 0.00597\tau_{13}^{p}\frac{\partial u}{\partial z} + 0.198\tau_{11}^{p}\tau_{11}^{p} + 0\tau_{12}^{p}\tau_{12}^{p} + 0\tau_{13}^{p}\tau_{13}^{p} + \cdots$ | 65.4% | 67 |
| | | $0.0043\dot{\gamma}_{12} = 0.720\tau_{12}^{p} + 0.00399u\frac{\partial \tau_{12}^{p}}{\partial x} + 0.00395v\frac{\partial \tau_{12}^{p}}{\partial y} + 0.00397w\frac{\partial \tau_{12}^{p}}{\partial z} - 0.00368\tau_{11}^{p}\frac{\partial v}{\partial x} - 0.00341\tau_{12}^{p}\frac{\partial v}{\partial y} - 0.00416\tau_{13}^{p}\frac{\partial v}{\partial z} - 0.00350\tau_{12}^{p}\frac{\partial u}{\partial x} - 0.00347\tau_{22}^{p}\frac{\partial u}{\partial y} - 0.00431\tau_{23}^{p}\frac{\partial u}{\partial z} + 0.402\tau_{11}^{p}\tau_{12}^{p} + 0.386\tau_{12}^{p}\tau_{22}^{p} + 0.506\tau_{13}^{p}\tau_{23}^{p} + \cdots$ | 50.6% | 39 |
| | | $0.0043\dot{\gamma}_{13} = 0.739\tau_{13}^{p} + 0.00284u\frac{\partial \tau_{13}^{p}}{\partial x} + 0.00273v\frac{\partial \tau_{13}^{p}}{\partial y} + 0.00244w\frac{\partial \tau_{13}^{p}}{\partial z} - 0.00133\tau_{11}^{p}\frac{\partial w}{\partial x} - 0.00333\tau_{12}^{p}\frac{\partial w}{\partial y} + 0\tau_{13}^{p}\frac{\partial w}{\partial z} + 0\tau_{13}^{p}\frac{\partial u}{\partial x} - 0.00223\tau_{23}^{p}\frac{\partial u}{\partial y} - 0.00185\tau_{33}^{p}\frac{\partial u}{\partial z} + 0\tau_{11}^{p}\tau_{13}^{p} + 0.230\tau_{12}^{p}\tau_{23}^{p} + 0\tau_{13}^{p}\tau_{33}^{p} + \cdots$ | 76.3% | 135 |
| | | $0.0043\dot{\gamma}_{22} = 0.447\tau_{22}^{p} + 0.00213u\frac{\partial \tau_{22}^{p}}{\partial x} + 0.00229v\frac{\partial \tau_{22}^{p}}{\partial y} + 0.00213w\frac{\partial \tau_{22}^{p}}{\partial z} - 0.00431\tau_{12}^{p}\frac{\partial v}{\partial x} + 0\tau_{22}^{p}\frac{\partial v}{\partial y} - 0.00411\tau_{23}^{p}\frac{\partial v}{\partial z} + 0\tau_{12}^{p}\tau_{12}^{p} + 0.238\tau_{22}^{p}\tau_{22}^{p} + 0\tau_{23}^{p}\tau_{23}^{p} + \cdots$ | 79.5% | 58 |
| | | $0.0043\dot{\gamma}_{23} = 0.423\tau_{23}^{p} + 0.00188u\frac{\partial \tau_{23}^{p}}{\partial x} + 0.00159v\frac{\partial \tau_{23}^{p}}{\partial y} + 0.00106w\frac{\partial \tau_{23}^{p}}{\partial z} - 0.00142\tau_{12}^{p}\frac{\partial w}{\partial x} - 0.000552\tau_{22}^{p}\frac{\partial w}{\partial y} - 0.000469\tau_{23}^{p}\frac{\partial w}{\partial z} - 0.000879\tau_{13}^{p}\frac{\partial v}{\partial x} + 0\tau_{23}^{p}\frac{\partial v}{\partial y} + 0\tau_{33}^{p}\frac{\partial v}{\partial z} + 0\tau_{12}^{p}\tau_{13}^{p} + 0\tau_{22}^{p}\tau_{23}^{p} + 0\tau_{23}^{p}\tau_{33}^{p} + \cdots$ | 89.2% | 41 |
| | | $0.0043\dot{\gamma}_{33} = 0.687\tau_{33}^{p} + 0.00185u\frac{\partial \tau_{33}^{p}}{\partial x} + 0.00167v\frac{\partial \tau_{33}^{p}}{\partial y} + 0.00184w\frac{\partial \tau_{33}^{p}}{\partial z} - 0.00322\tau_{13}^{p}\frac{\partial w}{\partial x} - 0.00358\tau_{23}^{p}\frac{\partial w}{\partial y} - 0.00233\tau_{33}^{p}\frac{\partial w}{\partial z} + 0\tau_{13}^{p}\tau_{13}^{p} + 0\tau_{23}^{p}\tau_{23}^{p} + 0\tau_{33}^{p}\tau_{33}^{p} + \cdots$ | 80.7% | 54 |

## IV. Discussion

### A. The role of Cartesian tensor

Cartesian tensors are fundamental to the CTSR method. Herein, we analyze how Cartesian tensors facilitate the accurate discovery of governing equations.

As detailed in Section II, the construction of the candidate library involves systematically eliminating candidates that fail to meet the criteria for validity and uniqueness. This reduction in candidate numbers significantly simplifies the discovery process, which represents a primary function of Cartesian tensors in achieving accurate identification.

Taking the 3D Navier-Stokes case as an example, Figure 9 illustrates the evolution of candidate coefficients during the sparse regression training process. For both CTSR and PDE-FIND, the coefficients corresponding to the exact governing equation emerge early in training. The primary role of TrainSTRidge is to eliminate redundant terms, thus promoting sparsity in the solution. Specifically, the second to fourth rows in Fig. 9 show that while a large number of candidate coefficients are nonzero at the start of training, TrainSTRidge progressively prunes most extraneous terms. With CTSR significantly curtailing the candidate pool, the challenge of obtaining an accurate and concise equation is markedly reduced, leading to enhanced prediction accuracy.

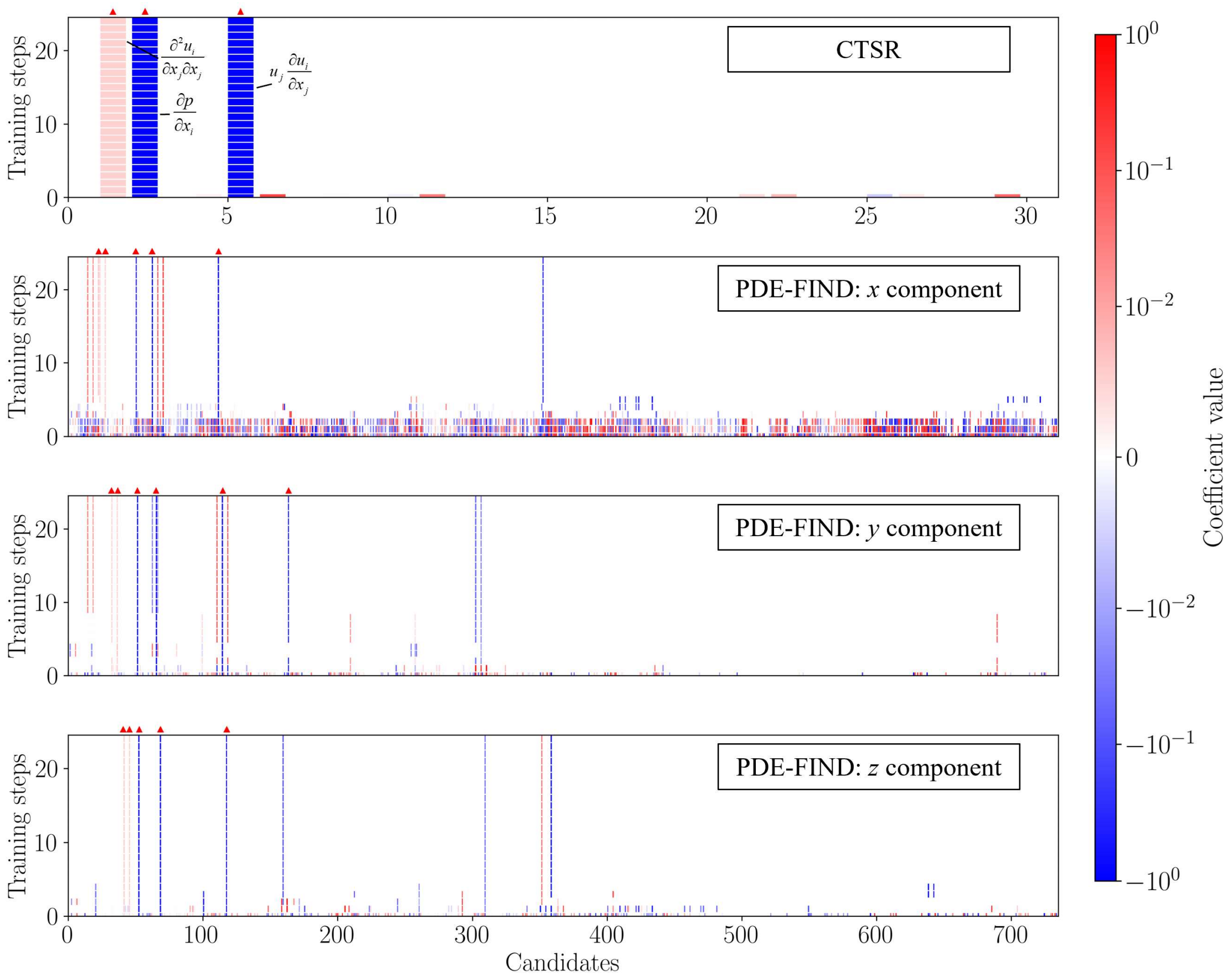

**Fig. 9.** Evolution of candidate coefficients during TrainSTRidge training iterations. The columns marked with red triangles indicate terms present in the exact equation.

Figure 9 also indicates that both CTSR and PDE-FIND achieve stable results before 10 training steps. This observation suggests that the selected number of training steps (i.e., $n_{\text{train}}$ = 25) is adequate and does not constrain algorithm performance. In the case of PDE-FIND, merely increasing $n_{\text{train}}$ cannot improve the accuracy.

The impact of the number of candidate terms on prediction error is further demonstrated through a second example. Figure 10 presents the candidate counts and average prediction errors for Case 2 under different *P*, using both CTSR and PDE-FIND. In this figure, bar graphs denote the number of candidates while lines represent the average errors averaged over multiple experiments. As *P* increases, the candidate count produced by PDE-FIND grows rapidly, and the error also shows an increasing trend. In contrast, although the candidate count for CTSR also increases with *P*, it remains markedly lower than that of PDE-FIND at the same *P* values, with its average error consistently below 1%.

Notably, even though PDE-FIND generates fewer candidate terms at *P* = 1 than CTSR does at *P* = 3 and 4, CTSR still achieves a lower error. This suggests that the enhanced accuracy of CTSR is not solely attributed to the reduced number of candidates; other contributing factors are also involved.

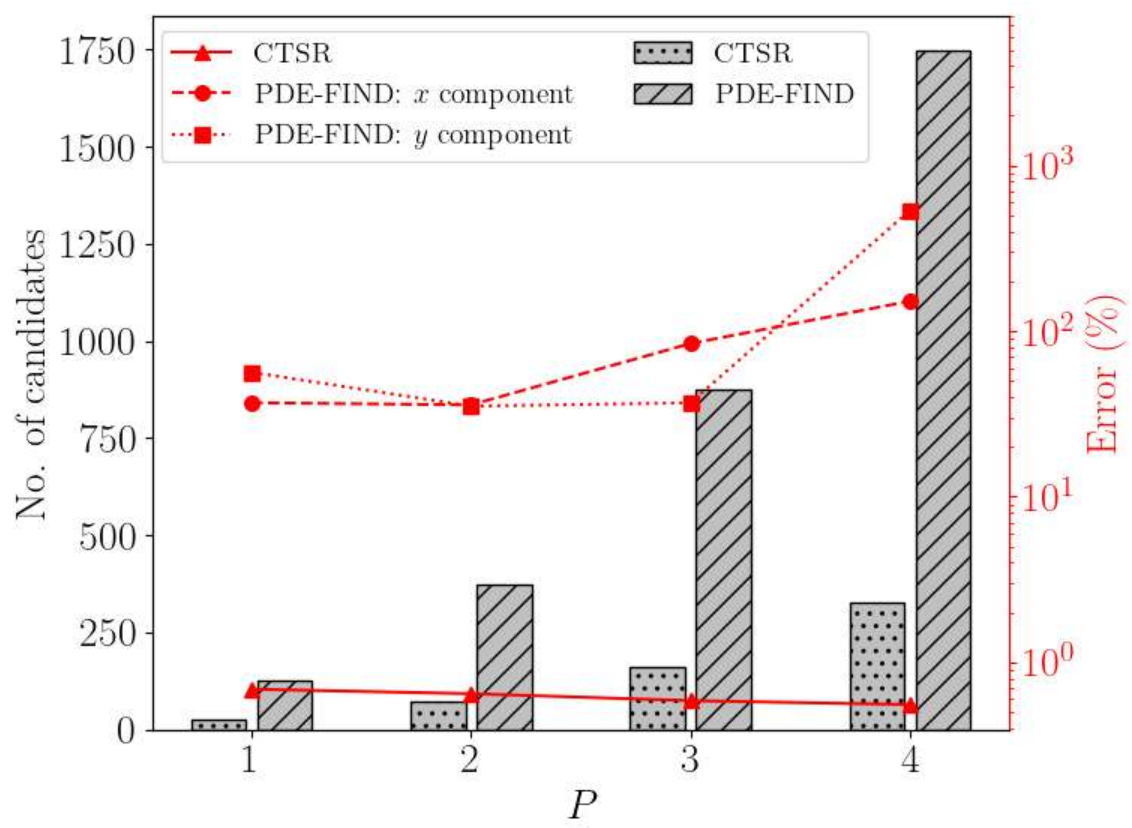

**Fig. 10.** Candidate counts and average prediction errors for Case 2 under different values of *P*.

As discussed in Section III, in Case 3, the sparse distribution of flow features around the cylinder along the *z*-direction results in poor accuracy when PDE-FIND predicts the *z*-component equation. To further analyze this phenomenon, the matrix $\mathbf{\Theta}$ and the vector $\mathbf{u}_t$, obtained via the CTSR method, are partitioned into three groups corresponding to the *x*, *y*, and *z* directions, followed by separate sparse regressions on each group. The resulting prediction errors, obtained from multiple experiments and compared against those of PDE-FIND, are summarized in Table VII. Here, we set *P* = 1 to highlight the influence of data quality across different dimensions. The “Reference” column in Table VII reports the error obtained by CTSR using the complete, undivided dataset.

Table VII shows that using PDE-FIND results in a markedly higher prediction error for the *z*-component compared to the *x* and *y* components. Specifically, the *z*-direction errors have a mean of 150.9% and a standard deviation of 182.7%, indicating a significant failure in accurately identifying the corresponding equation. Similarly, when CTSR is applied exclusively to *z*-direction data, the observed errors are significantly higher than those obtained from the other two dimensions, highlighting the critical influence of data quality on accuracy. In contrast, when the CTSR method is applied to the complete, undivided dataset, it achieves a mean error of only 0.70% with a standard deviation of 0.49%. This improvement is attributed to the coupling of multi-dimensional data within the constructed matrix $\mathbf{\Theta}$ and vector $\mathbf{u}_t$, which mitigates the detrimental effects of poor-quality *z*-direction data. Fundamentally, this benefit arises because the form of the Cartesian tensor equation is invariant across different dimensions, enabling the simultaneous use of data from all dimensions. Thus, another key function of Cartesian tensors is their ability to couple multi-dimensional data, which enhances the accuracy.

**Table VII.** Prediction errors obtained using data from different dimensions, compared with the PDE-FIND results ($P$ = 1).

| | *x* part/component | *y* part/component | *z* part/component | **Reference** |
|---|---|---|---|---|
| CTSR | 1.49±0.76% | 0.94±0.53% | 15.8±9.10% | **0.70±0.49%** |
| PDE-FIND | 46.7±12.2% | 51.6±33.4% | 150.9±182.7% | \ |

Similar scenarios to Case 3 frequently arise in practical applications, where data in certain directions may suffer from insufficient sampling, high noise levels, or significant systematic errors. In such situations, the CTSR method is expected to deliver better accuracy compared to conventional scalar-based approaches.

## B. Effect of sampling points

In Section III, the datasets used for regression are generated by randomly sub-sampling from the spatial and temporal domains using a fixed random seed (i.e., a fixed distribution). Here, different random seeds are employed to adjust the space–time distribution of the sampling points, and multiple experiments are conducted to evaluate the algorithm's robustness. We also analyzed the prediction error as a function of the number of sampling points. The hyperparameter and $P$ remain consistent with those in Section III. Figure 11 presents the results: blue boxplots depict the error distribution of CTSR, and the solid lines show the average prediction errors for CTSR and PDE-FIND.

For all four cases investigated, the CTSR method consistently yields a lower average error than PDE-FIND. With the exception of the relatively simple Case 1, the prediction error of CTSR is approximately one to two orders of magnitude lower than that of PDE-FIND. Moreover, sampling density affects the average error: as density increases, the average error decreases slightly before stabilizing.

Sampling density also impacts the error distribution. In Case 2 and 4, at low sampling densities, errors are widely dispersed; however, as density increases, the error distribution becomes more concentrated, and the average error converges to a stable value. Notably, at low sampling density, outlier points with errors significantly above and below the eventual stable average arise. This phenomenon stems from the strong randomness in sample coverage at low densities, which introduces considerable uncertainty in feature capture. When sampling points cluster in regions rich in flow features, the prediction error tends to be low; conversely, if most samples fall within regions with relatively homogeneous flow features, the accuracy can deteriorate markedly. As sampling density increases, broader coverage induces statistical stability in the error distribution, rather than reaching its minimum. In Case 1 and 3, this phenomenon is less pronounced due to the more uniform distribution of flow features.

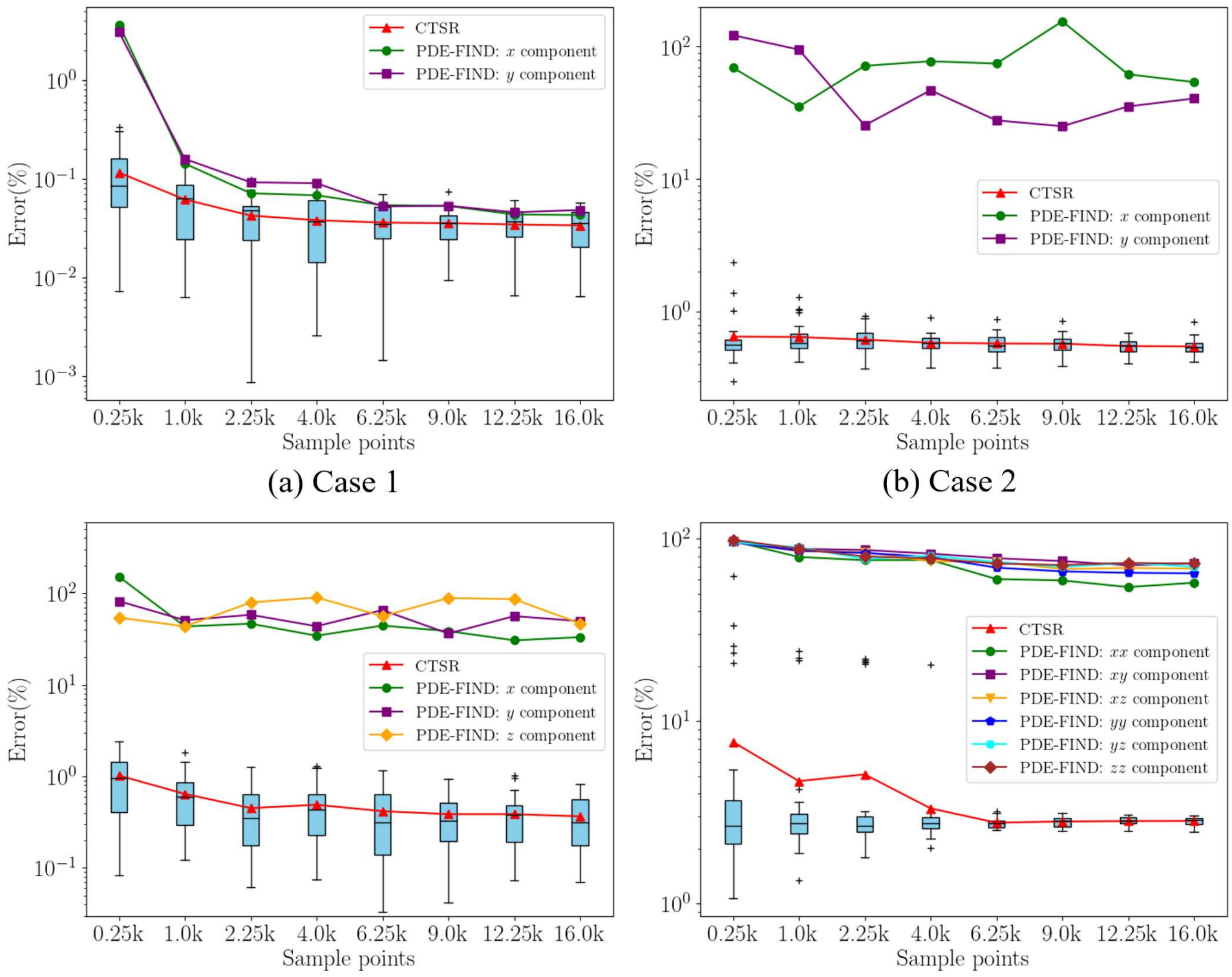

(a) Case 1 (b) Case 2

(c) Case 3 (d) Case 4

**Fig. 11.** Effect of sampling points on prediction error.

In practice, the spatial distribution of data samples can be highly non-uniform. For instance, in Particle Image Velocimetry (PIV) experiments, measurements are extracted from camera-captured images that typically cover only portions of the flow field. Cases 1 and 2 are used as examples to evaluate the performance of CTSR under highly non-uniform sampling conditions. The original spatial domain ($H \times H$) is divided into 16 approximately equal sampling windows, arranged in a 4×4 grid. Spatial sample points are randomly selected within each window, along with random sampling over the temporal domain. The total number of spatial and temporal sampling points remains consistent with the settings described in Section III. The resulting prediction errors, averaged over multiple trials, are presented in Fig. 12.

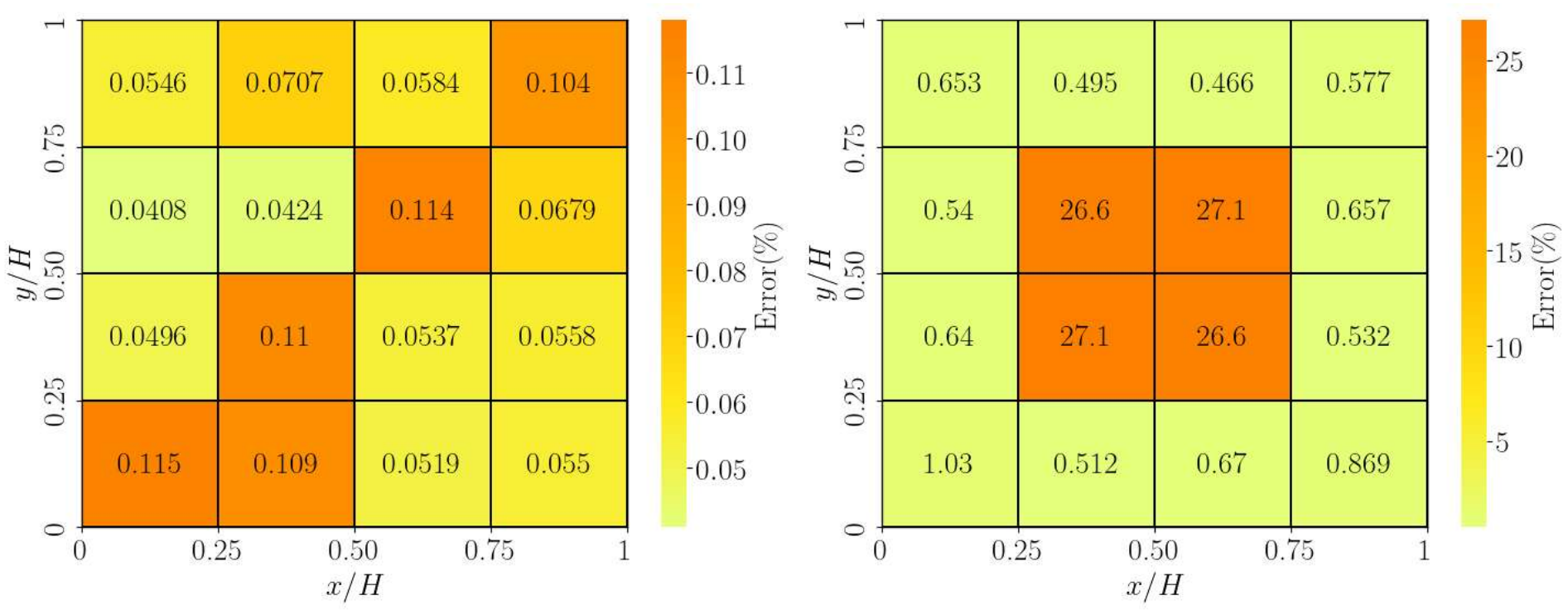

(a) Case 1 (b) Case 2

**Fig. 12.** Prediction errors across different spatial sampling windows.

As illustrated in Fig. 12, CTSR achieves consistently low prediction errors across all subregions in Case 1, with a standard deviation of approximately 0.027%. In contrast, Case 2 shows significant spatial variation: regions near the wall exhibit errors around 0.6%, while errors in regions farther from the wall can reach up to 27%, resulting in a standard deviation of approximately 11.4%. This discrepancy can be attributed to differences in the spatial distribution of flow features between the two cases. Case 1 involves a synthetically generated vorticity field with spatially uniform properties (see Fig. 4), which leads to relatively uniform prediction accuracy. In Case 2, however, the flow is driven by natural convection, with dominant structures concentrated near the wall (see Fig. 5). As a result, prediction accuracy is higher in regions adjacent to the wall, whereas regions farther away exhibit increased errors.

Overall, CTSR demonstrates robust performance across different sampling densities; however, its accuracy is notably influenced by the richness of the data features, which in turn depends on the distribution of sampling points. In practical applications — especially when faced with sparse data or inhomogeneous feature distributions — it is advisable to prioritize data acquisition in regions with more pronounced flow features. Such a strategy can help enhance the reliability and accuracy of equation discovery.

### C. Runtime performance

Sparse regression techniques are valued for their efficiency in discovering governing equations—a virtue that should not be compromised by the introduction of Cartesian tensors. To assess the computational efficiency of CTSR, Table VIII presents a quantitative comparison of the runtimes for both the CTSR and PDE-FIND methods across four test cases. The experiments are conducted using an AMD 7950X processor with 64GB of RAM. The total runtime is divided into two stages—candidate library construction and sparse regression—with the reported values representing averages from multiple experiments.

As indicated in Table VIII, while CTSR requires slightly more time for candidate library construction in most cases—primarily due to the additional computations involved in generating the candidate library—it benefits from producing a smaller number of candidates, resulting in a significantly faster sparse regression stage compared to PDE-FIND. Overall, the CTSR method demonstrates a substantial runtime advantage over PDE-FIND, particularly in complex, high-dimensional problems (e.g., Cases 2 to 4).

**Table VIII.** Runtime of the CTSR and PDE-FIND

| | | Candidate library construction | Sparse regression | Acceleration |
|---|---|---|---|---|
| Case 1 | CTSR | ~0.02s | ~0.02s | ~4 |
| | PDE-FIND | ~0.002s | ~0.15s | |
| Case 2 | CTSR | ~0.03s | ~0.06s | ~32 |
| | PDE-FIND | ~0.02s | ~2.9s | |
| Case 3 | CTSR | ~0.02s | ~0.04s | ~161 |
| | PDE-FIND | ~0.05s | ~9.6s | |
| Case 4 | CTSR | ~1.6s | ~0.3s | ~59 |
| | PDE-FIND | ~0.3s | ~112s | |

## V. Conclusion

This paper introduces CTSR, a Cartesian tensor-based sparse regression technique for discovering governing equations in high-dimensional dynamical systems. By employing Cartesian tensor equations as the target form and constructing a tensor candidate library, CTSR accurately discovers complex, high-dimensional governing equations while ensuring invariance. The method was validated using four test cases of increasing complexity and dimensionality. Results indicate that CTSR accurately discovers all the equations and produces lower errors than PDE-FIND, with its performance advantages becoming increasingly pronounced in higher-dimensional problems. Furthermore, an analysis of the influence of sampling point density and distribution reveals CTSR's robust performance with respect to sampling density and its sensitivity to the richness of data features. Lastly, an evaluation of runtime performance confirms CTSR's ability to efficiently discover equations.

The proposed CTSR method exhibits several key advantages that contribute to its considerable performance. First, the invariance property of the Cartesian tensor under coordinate rotations and reflections enables the discovery of invariant governing equations. At the same time, employing Cartesian tensors reduces the number of candidate terms, thereby reducing the difficulty of identifying the correct equations, particularly in complex, high-dimensional settings. Second, unlike scalar expressions, Cartesian tensor equations maintain the same form across different dimensions. This characteristic promotes effective data coupling among spatial dimensions and mitigates issues arising from poor data quality in any individual dimension.

It is important to acknowledge some limitations of CTSR. First, as with many sparse regression methods, its representational capacity is constrained by predefined assumptions regarding the form of the target equations. This constraint can inhibit the discovery of certain equation types, such as fractional-order partial differential equations. Second, CTSR is sensitive to data noise — a common challenge in sparse regression approaches — which may limit its applicability to experimental or observational datasets characterized by high noise levels. Potential mitigation strategies include preprocessing techniques such as Singular Value Decomposition (SVD) for noise reduction, or employing more robust derivative estimation methods, such as differentiation based on Savitzky-Golay filtering. Recent studies [23,56] have also shown that formulating the equations in their weak form can substantially improve robustness to noise. Finally, the high-dimensional nature of Cartesian tensors can lead to non-unique representations in low-dimensional scenarios. For instance, in a 1D system along the $x$-axis, expressions such as $u_i \partial u_j / \partial x_j$, $u_j \partial u_i / \partial x_j$, and $u_j \partial u_j / \partial x_i$ all reduce to $u \partial u / \partial x$, which increases the complexity of discovering governing equations of low-dimensional systems. While this does not pose a fundamental obstacle to equation discovery, it may lead to redundant candidate terms. This issue is primarily confined to 1D settings, which lie outside the main scope of CTSR. When necessary, these redundancies can be resolved by eliminating duplicate columns in the matrix $\mathbf{\Theta}$.

Several factors that may influence the accuracy of CTSR warrant further discussion. One key factor is the numerical scheme used for derivative estimation. In this study, second-order finite-difference methods (Cases 1–3) and cell-based derivative [55] (Case 4) were employed. While higher-order schemes can improve accuracy in flows with fine-scale structures (e.g., small vortices) by reducing truncation errors, their greater sensitivity to noise and implementation difficulty on unstructured meshes may limit their practical applicability. Another factor is the boundary artifacts introduced by inaccurate derivative estimations near domain edges. To mitigate this, data at boundary nodes were excluded from sampling in this study. While effective, this approach may omit important physical information — such as wall shear stress or heat flux — which can be critical in certain applications. In such cases, more careful boundary treatments or higher-order discretization may be required to preserve accuracy without discarding valuable data.

Although sparse regression serves as the primary methodological foundation of CTSR, it is not the only viable framework for equation discovery. The core principle of employing Cartesian tensors can be readily integrated into other state-of-the-art techniques, offering the potential to enhance their performance in high-dimensional settings while preserving their inherent strengths. Future enhancements to CTSR will integrate physical constraints (e.g., dimensional consistency) and improve robustness against data noise. Moreover, while the present work focuses on validating the proposed method, future research will apply it to constitutive or closure modeling challenges in complex, high-dimensional dynamical systems.

## Appendix A: Selection of $d_{\mathrm{tol}}$

Within the CTSR method, the initial tolerance, $d_{\mathrm{tol}}$, substantially influences the accuracy of the discovered equations. We employ a Pareto front analysis to determine an optimal $d_{\mathrm{tol}}$ value that strikes a balance between accuracy and conciseness in the resulting equations.

For all four cases studied, the objectives are defined as equation sparsity (characterized by $\|\hat{\boldsymbol{\xi}}\|_0$) and error (characterized by $\|\mathbf{u}_t - \boldsymbol{\Theta}\hat{\boldsymbol{\xi}}\|_2$), with $d_{\mathrm{tol}}$ serving as the optimization variable. Through a Pareto front analysis, $d_{\mathrm{tol}}$ is varied over a broad range from $10^{-5}$ to $10^{3}$, and the resulting $\|\hat{\boldsymbol{\xi}}\|_0$ and $\|\mathbf{u}_t - \boldsymbol{\Theta}\hat{\boldsymbol{\xi}}\|_2$ are recorded and presented in a scatter diagram (see Fig. A.1). In this figure, each triangle represents a prediction (with overlap among many points), and its color indicates the $d_{\mathrm{tol}}$ value. Dashed lines and arrows indicate the Pareto front and the general direction of increasing $d_{\mathrm{tol}}$, respectively. Red circles highlight the instances where the correct equations were discovered; $d_{\mathrm{tol}}$ values from these instances are used in the main text.

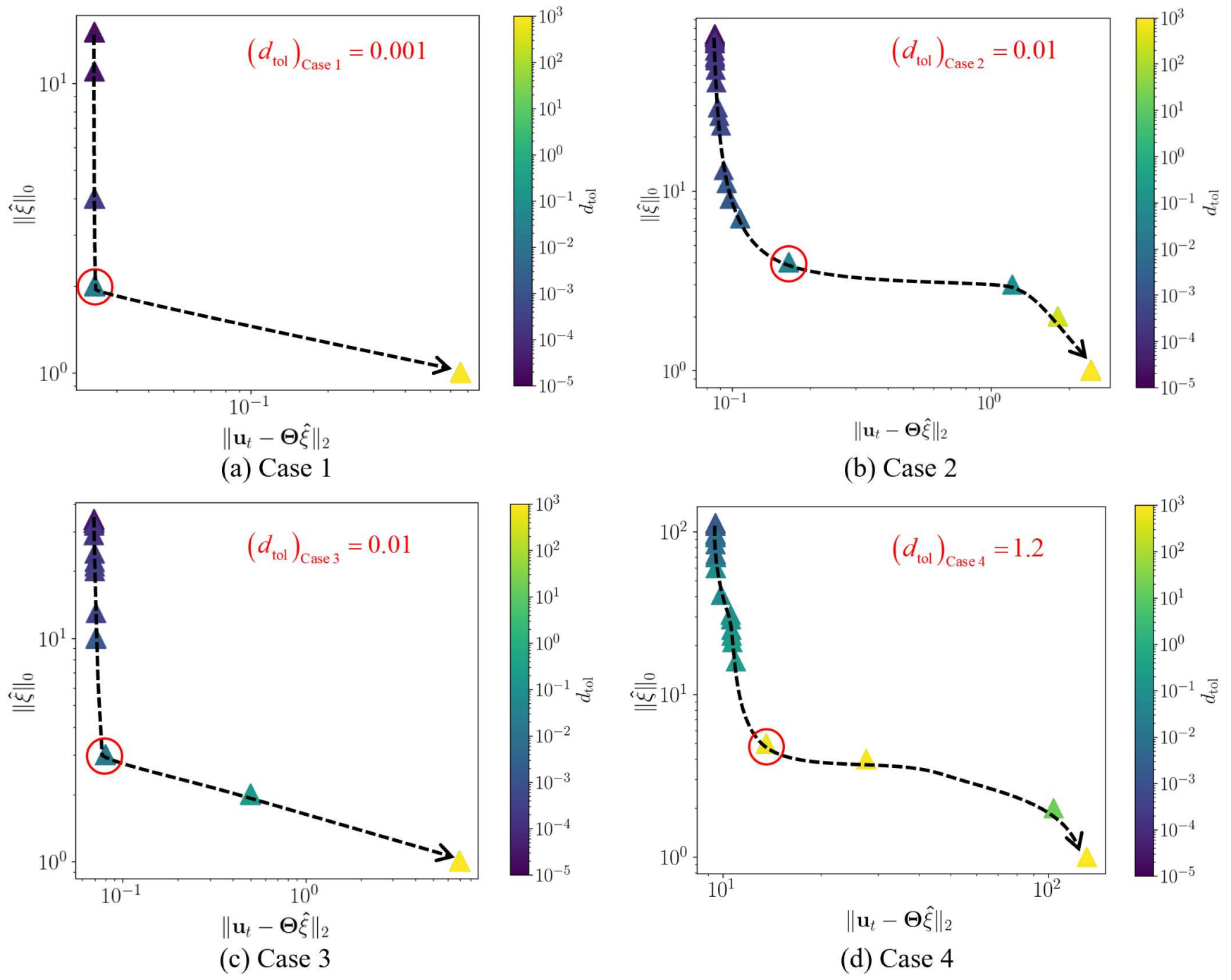

**Fig. A.1.** Pareto front analysis results for the four test cases

As illustrated in Fig. A.1, $d_{tol}$ significantly affects both the sparsity and error of the discovered equations. In general, increasing $d_{tol}$ results in more concise equations (i.e., lower $\|\hat{\boldsymbol{\xi}}\|_0$), while simultaneously increasing the error $\|\mathbf{u}_t - \boldsymbol{\Theta}\hat{\boldsymbol{\xi}}\|_2$. Specifically, selecting a small $d_{tol}$ yields very low error values but introduces many extraneous terms, some of which may originate from data noise or inaccuracies in numerical differentiation. Conversely, an excessively large $d_{tol}$ produces oversimplified predictions with substantially higher error.

The red-circled points denote the $d_{tol}$ value at which CTSR accurately discovered the governing equations. These points correspond to inflection points near the lower left corner of the plotted curves, where the predicted equations achieve a balance between conciseness and accuracy. Therefore, in practical applications, an optimal $d_{tol}$ can be determined by identifying these inflection points.

## Declaration of competing interest

The authors have no conflicts to disclose.

## Acknowledgments

The authors would like to thank Prof. Richard D. Sandberg and Maximilian Reissmann of the University of Melbourne for kindly sharing their code, which served as a helpful reference in the development of this work.

## Data Availability

Data will be made available on request.

## References

[1] S.L. Brunton, and J.N. Kutz, "Promising directions of machine learning for partial differential equations," Nat Comput Sci, (2024).
[2] N. Makke, and S. Chawla, "Interpretable scientific discovery with symbolic regression: a review," Artif Intell Rev **57**(1), 2 (2024).
[3] JohnR. Koza, "Genetic programming as a means for programming computers by natural selection," Stat Comput **4**(2), (1994).
[4] J. Bongard, and H. Lipson, "Automated reverse engineering of nonlinear dynamical systems," Proc. Natl. Acad. Sci. U.S.A. **104**(24), 9943–9948 (2007).
[5] M. Schmidt, and H. Lipson, "Distilling free-form natural laws from experimental data," Science **324**(5923), 81–85 (2009).
[6] E.J. Vladislavleva, G.F. Smits, and D. Den Hertog, "Order of nonlinearity as a complexity measure for models generated by symbolic regression via pareto genetic programming," IEEE Trans. Evol. Comput. **13**(2), 333–349 (2009).
[7] R. Dubčáková, "Eureqa: software review," Genet. Prog. Evolv. Mach. **12**(2), 173–178 (2011).
[8] M. Virgolin, T. Alderliesten, C. Witteveen, and P.A.N. Bosman, "Improving Model-based Genetic Programming for Symbolic Regression of Small Expressions," Evol. Comput. **29**(2), 211–237 (2021).
[9] Y. Chen, Y. Luo, Q. Liu, H. Xu, and D. Zhang, "Symbolic genetic algorithm for discovering open-form partial differential equations (SGA-PDE)," Phys. Rev. Res. **4**(2), 023174 (2022).
[10] W. Ma, J. Zhang, K. Feng, H. Xing, and D. Wen, "Dimensional homogeneity constrained gene expression programming for discovering governing equations," J. Fluid Mech. **985**, A12 (2024).
[11] M. Cranmer, "Interpretable Machine Learning for Science with PySR and SymbolicRegression.jl," (2023).
[12] M. Reissmann, Y. Fang, A.S.H. Ooi, and R.D. Sandberg, "Constraining genetic symbolic regression via semantic backpropagation," Genet. Prog. Evolv. Mach. **26**(1), 12 (2025).
[13] G. Martius, and C.H. Lampert, "Extrapolation and learning equations," (2016).
[14] S.S. Sahoo, C.H. Lampert, and G. Martius, "Learning equations for extrapolation and control," (PMLR, 2018), pp. 4442–4450.
[15] Z. Long, Y. Lu, and B. Dong, "PDE-Net 2.0: learning PDEs from data with a numeric-symbolic hybrid deep network," J. Comput. Phys. **399**, 108925 (2019).
[16] S.-M. Udrescu, and M. Tegmark, "AI Feynman: A physics-inspired method for symbolic regression," Sci. Adv. **6**(16), (2020).
[17] S. Kim, P.Y. Lu, S. Mukherjee, M. Gilbert, L. Jing, V. Ceperic, and M. Soljacic, "Integration of neural network-based symbolic regression in deep learning for scientific discovery," IEEE Trans. Neural Netw. Learning Syst. **32**(9), 4166–4177 (2021).
[18] B.K. Petersen, M. Landajuela, T.N. Mundhenk, C.P. Santiago, S.K. Kim, and J.T. Kim, "Deep symbolic regression: Recovering mathematical expressions from data via risk-seeking policy gradients," (2021).
[19] M. Du, Y. Chen, and D. Zhang, "DISCOVER: Deep identification of symbolically concise open-form partial differential equations via enhanced reinforcement learning," Phys. Rev. Res. **6**(1), 013182 (2024).
[20] G. Norman, J. Wentz, H. Kolla, K. Maute, and A. Doostan, "Constrained or unconstrained? Neural-network-based equation discovery from data," Comput. Method Appl. Mech. Eng. **436**, 117684 (2025).

[21] S.L. Brunton, J.L. Proctor, and J.N. Kutz, "Discovering governing equations from data by sparse identification of nonlinear dynamical systems," Proc. Natl. Acad. Sci. U.S.A. **113**(15), 3932–3937 (2016).
[22] S.H. Rudy, S.L. Brunton, J.L. Proctor, and J.N. Kutz, "Data-driven discovery of partial differential equations," Sci. Adv. **3**(4), e1602614 (2017).
[23] D.R. Gurevich, P.A.K. Reinbold, and R.O. Grigoriev, "Robust and optimal sparse regression for nonlinear PDE models," Chaos **29**(10), 103113 (2019).
[24] P.A.K. Reinbold, "Robust learning from noisy, incomplete, high-dimensional experimental data via physically constrained symbolic regression," Nat Commun **12**(1), (2021).
[25] K. Champion, B. Lusch, J.N. Kutz, and S.L. Brunton, "Data-driven discovery of coordinates and governing equations," Proc. Natl. Acad. Sci. U.S.A. **116**(45), 22445–22451 (2019).
[26] Y. Li, K. Wu, and J. Liu, "Discover governing differential equations from evolving systems," Phys. Rev. Res. **5**(2), 023126 (2023).
[27] Z. Chen, Y. Liu, and H. Sun, "Physics-informed learning of governing equations from scarce data," Nat Commun **12**(1), 6136 (2021).
[28] P.Y. Lu, J. Ariño Bernad, and M. Soljačić, "Discovering sparse interpretable dynamics from partial observations," Commun Phys **5**(1), 206 (2022).
[29] K. Kaheman, J.N. Kutz, and S.L. Brunton, "SINDy-PI: a robust algorithm for parallel implicit sparse identification of nonlinear dynamics," Proc. R. Soc. A. **476**(2242), 20200279 (2020).
[30] J. Wentz, and A. Doostan, "Derivative-based SINDy (DSINDy): Addressing the challenge of discovering governing equations from noisy data," Comput. Method Appl. Mech. Eng. **413**, 116096 (2023).
[31] R. Yu, and R. Wang, "Learning dynamical systems from data: An introduction to physics-guided deep learning," Proc. Natl. Acad. Sci. U.S.A. **121**(27), e2311808121 (2024).
[32] C. Chen, H. Li, and X. Jin, "An invariance constrained deep learning network for partial differential equation discovery," Phys. Fluids, (2024).
[33] X. Xie, A. Samaei, J. Guo, W.K. Liu, and Z. Gan, "Data-driven discovery of dimensionless numbers and governing laws from scarce measurements," Nat Commun **13**(1), 7562 (2022).
[34] J. Zhang, and W. Ma, "Data-driven discovery of governing equations for fluid dynamics based on molecular simulation," J. Fluid Mech. **892**, A5 (2020).
[35] C.M. Oishi, A.A. Kaptanoglu, J.N. Kutz, and S.L. Brunton, "Nonlinear parametric models of viscoelastic fluid flows," R. Soc. Open Sci. **11**(10), 240995 (2024).
[36] J. Weatheritt, and R. Sandberg, "A novel evolutionary algorithm applied to algebraic modifications of the RANS stress–strain relationship," J. Comput. Phys. **325**, 22–37 (2016).
[37] J. Wang, Y. Wang, H. Zhang, Z. Yang, Z. Liang, J. Shi, H.-T. Wang, D. Xing, and J. Sun, "E(n)-Equivariant cartesian tensor message passing interatomic potential," Nat Commun **15**(1), 7607 (2024).
[38] S.B. Pope, *Turbulent Flows* (Cambridge University Press, Cambridge ; New York, 2000).
[39] H. Schaeffer, "Learning partial differential equations via data discovery and sparse optimization," Proc. R. Soc. A. **473**(2197), 20160446 (2017).
[40] B. Zhao, M. He, and J. Wang, "Data-driven discovery of the governing equation of granular flow in the homogeneous cooling state using sparse regression," Phys. Fluids **35**(1), 013315 (2023).
[41] K. Raviprakash, B. Huang, and V. Prasad, "A hybrid modelling approach to model process dynamics by the discovery of a system of partial differential equations," Comput. Chem. Eng. **164**, 107862 (2022).
[42] M. Zhang, S. Kim, P.Y. Lu, and M. Soljačić, "Deep Learning and Symbolic Regression for Discovering Parametric Equations," IEEE Trans. Neural Netw. Learning Syst., 1–13 (2024).
[43] R. Stephany, and C. Earls, "PDE-LEARN: Using deep learning to discover partial differential equations from noisy, limited data," Neural Netw. **174**, 106242 (2024).
[44] P. Bartholomew, G. Deskos, R.A.S. Frantz, F.N. Schuch, E. Lamballais, and S. Laizet, "Xcompact3D: An open-source framework for solving turbulence problems on a Cartesian mesh," SoftwareX **12**, 100550 (2020).
[45] Z. Zhang, and Y. Liu, "A robust framework for identification of PDEs from noisy data," J. Comput. Phys. **446**, 110657 (2021).
[46] F. Sebilleau, R. Issa, S. Lardeau, and S.P. Walker, "Direct numerical simulation of an air-filled differentially heated square cavity with Rayleigh numbers up to 10 11," Int. J. Heat Mass Transf. **123**, 297–319 (2018).
[47] J. Zhang, and C. Dalton, "A three-dimensional simulation of a steady approach flow past a circular cylinder at low Reynolds number," Int. J. Numer. Meth. Fluids **26**(9), 1003–1022 (1998).
[48] L. Campo-Deaño, M.S.N. Oliveira, and F.T. Pinho, "A Review of Computational Hemodynamics in Middle Cerebral Aneurysms and Rheological Models for Blood Flow," Appl. Mech. Rev. **67**(3), 030801 (2015).

[49] J.G. Oldroyd, “On the formulation of rheological equations of state,” Proc. R. Soc. Lond. A **200**(1063), 523–541 (1950).
[50] H. Giesekus, “A simple constitutive equation for polymer fluids based on the concept of deformation-dependent tensorial mobility,” J. Non-Newton. Fluid Mech. **11**(1–2), 69–109 (1982).
[51] N.P. Thien, and R.I. Tanner, “A new constitutive equation derived from network theory,” J. Non-Newton. Fluid Mech. **2**(4), 353–365 (1977).
[52] F. Pimenta, and M.A. Alves, “Stabilization of an open-source finite-volume solver for viscoelastic fluid flows,” J. Non-Newton. Fluid Mech. **239**, 85–104 (2017).
[53] A. Shariatkhah, M. Norouzi, and M.R.H. Nobari, “Numerical simulation of blood flow through a capillary using a non-linear viscoelastic model,” Clin. Hemorheol. Microcirc. **62**(2), 109–121 (2016).
[54] M. Song, S. Wang, Q. Qian, Y. Zhou, Y. Luo, and X. Gong, “Intracranial aneurysm CTA images and 3D models dataset with clinical morphological and hemodynamic data,” Sci Data **11**(1), 1213 (2024).
[55] S. Will, M. Ken, and L. Bill, *The Visualization Toolkit (4th Ed.)* (Kitware, 2006).
[56] P.A.K. Reinbold, D.R. Gurevich, and R.O. Grigoriev, “Using noisy or incomplete data to discover models of spatiotemporal dynamics,” Phys. Rev. E **101**(1), 010203 (2020).